\documentclass[letterpaper, 10 pt, conference]{ieeeconf}  

\IEEEoverridecommandlockouts                              

\usepackage{textcase} 
\usepackage{graphics}
\usepackage{amsmath}
\usepackage{multirow}
\usepackage{makecell}
\usepackage[font=footnotesize,labelfont=bf]{caption}
\usepackage{subcaption}
\usepackage{rotating}
\usepackage{color}

\usepackage{afterpage}
\usepackage{float}
\usepackage{colortbl}
\usepackage{cuted}
\usepackage{xcolor}
\colorlet{light}{green!50}
\colorlet{lightlight}{green!25}
\newcommand{\fst}{\cellcolor{light}\bf}
\newcommand{\second}{\cellcolor{lightlight}}
\colorlet{colorSep}{blue!5}  
\usepackage{mathtools}
\usepackage{amssymb}

\title{\LARGE \bf
TivNe-SLAM: Dynamic Mapping and Tracking via Time-Varying Neural Radiance Fields
}

\author{Chengyao Duan$^{1}$ and Zhiliu Yang$^{1}$$^{,\;2}$$^{\;*}$ 
\thanks{This work was supported by the Program of Yunnan Key Laboratory of Intelligent Systems and Computing under Grant 202405AV340009, and Yunnan Science Foundation of Yunnan Provincial Department of Science and Technology under Grant 202301AU070200.}
\thanks{$^{1}$ School of Information Science and Engineering, Yunnan University, Kunming, Yunnan 650500, China.}
\thanks{$^{2}$ Yunnan Key Laboratory of Intelligent Systems and Computing, Yunnan University, Kunming, Yunnan 650500, China.} 
\thanks{$^*$ Corresponding author, \texttt{zhiliu.yang@ynu.edu.cn}}
}

\begin{document}

\maketitle
\begin{strip}
\vspace{-6em}
\begin{center}

\centering
\setlength{\tabcolsep}{0.1em}
{\renewcommand{\arraystretch}{1.5}
\begin{tabular}{c c c c }
\includegraphics[width=0.24\linewidth]{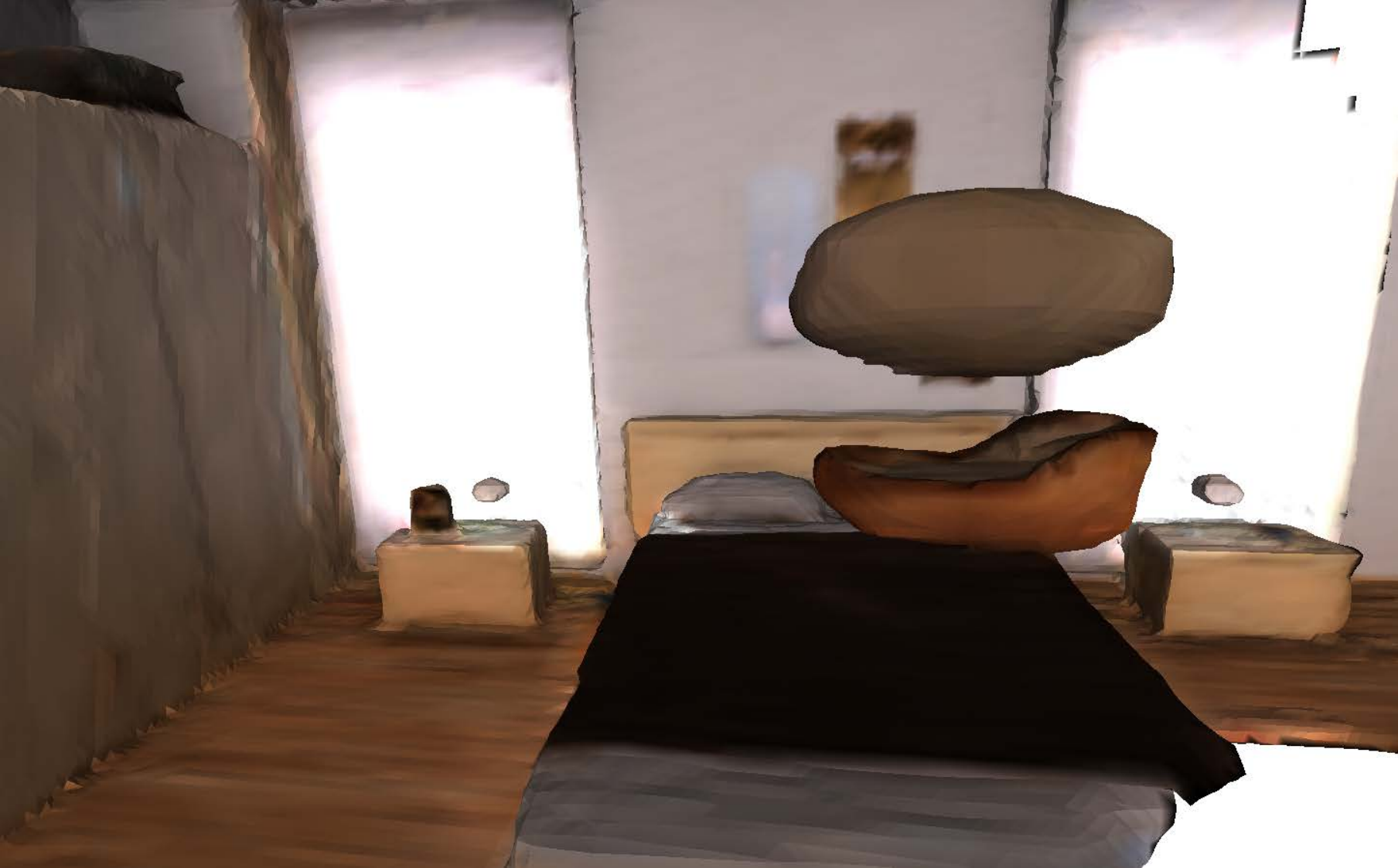} &
\includegraphics[width=0.24\linewidth]{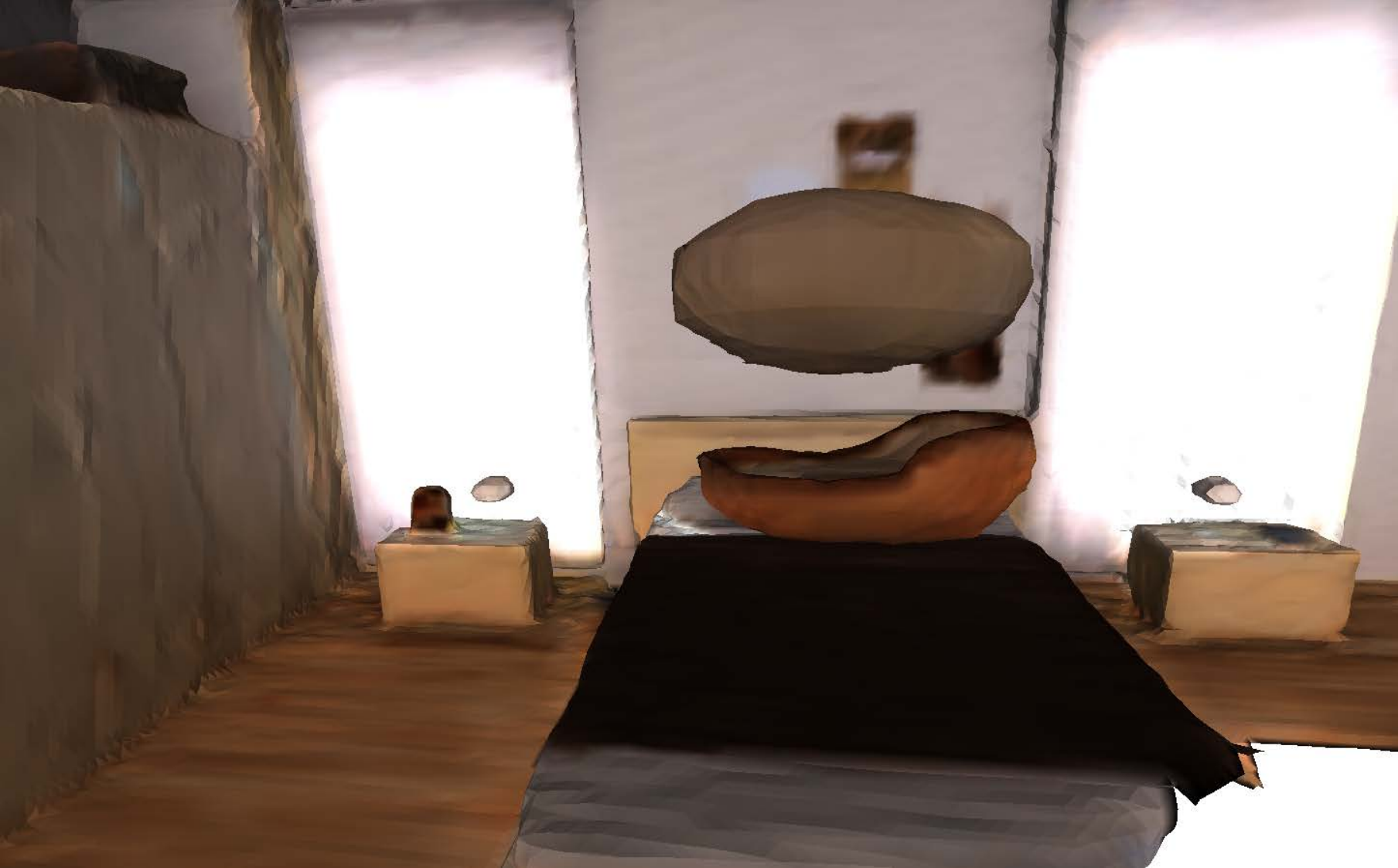} & 
\includegraphics[width=0.24\linewidth]{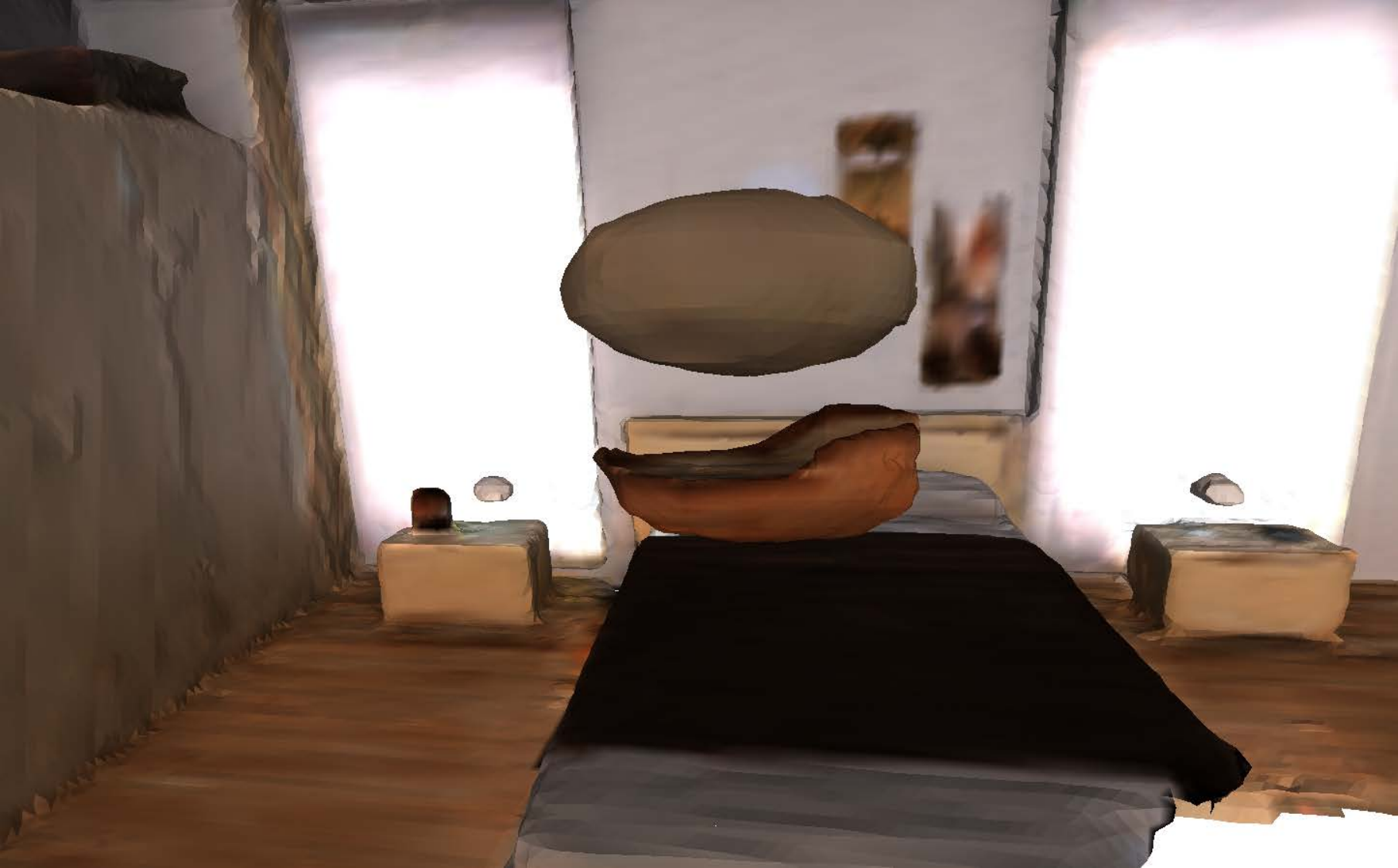}&
\includegraphics[width=0.24\linewidth]{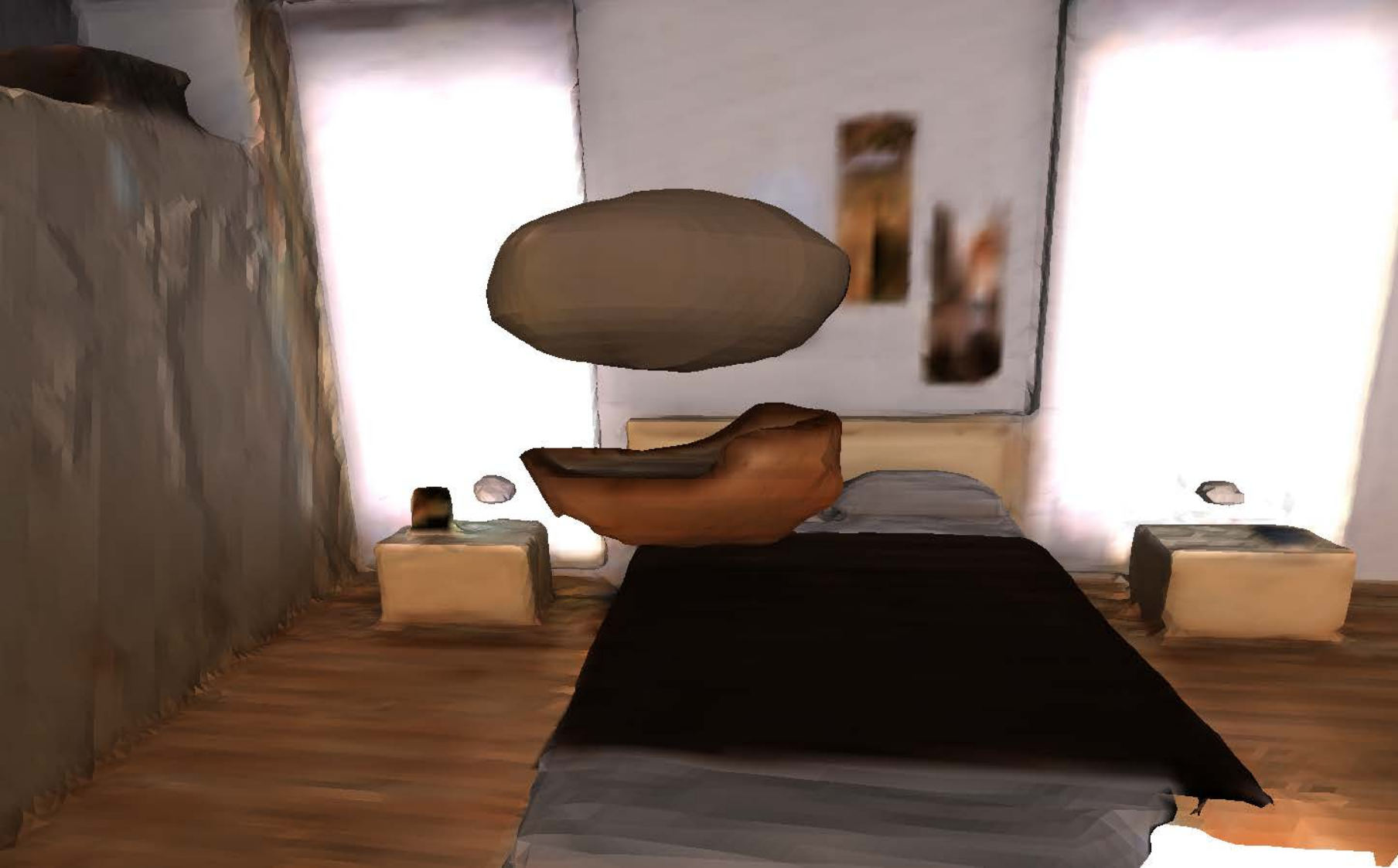}\vspace{-5pt}\\
\footnotesize(a) t = 22.96 s & \footnotesize(b) t = 23.56 s & \footnotesize(c) t = 24.16 s & \footnotesize(d) t = 24.76 s
\end{tabular}}
\vspace{-5pt}
\footnotesize

\captionof{figure}{\textbf{Illustration of the reconstruction of a dynamic object from our TivNe-SLAM.} We introduce a dynamic SLAM system capable of camera tracking and reconstruction of moving objects. Here, we show our 3D reconstruction result, at different time stamps, of a scene from the Room4 dataset \cite{runz2017co}, a flying ship is moving from the right to the left side of the room. The status of the flying ship is successfully captured and precisely reconstructed.}
\label{fig:guidance}
\vspace{-0.7cm}

\end{center}
\end{strip}
\thispagestyle{empty}
\pagestyle{empty}
\begin{abstract}

Previous attempts to integrate Neural Radiance Fields (NeRF) into the Simultaneous Localization and Mapping (SLAM) framework either rely on the assumption of static scenes or require the ground truth camera poses, which impedes their application in real-world scenarios. This paper proposes a time-varying representation to track and reconstruct the dynamic scenes. Firstly, two processes, a tracking process and a mapping process, are maintained simultaneously in our framework. In the tracking process, all input images are uniformly sampled and then progressively trained in a self-supervised paradigm. In the mapping process, we leverage motion masks to distinguish dynamic objects from the static background, and sample more pixels from dynamic areas. Secondly, the parameter optimization for both processes is comprised of two stages: the first stage associates time with 3D positions to convert the deformation field to the canonical field. The second stage associates time with the embeddings of the canonical field to obtain colors and a Signed Distance Function (SDF). Lastly, we propose a novel keyframe selection strategy based on the overlapping rate. Our approach is evaluated on two synthetic datasets and one real-world dataset, and the experiments validate that our method achieves competitive results in both tracking and mapping when compared to existing state-of-the-art NeRF-based dynamic SLAM systems.
\end{abstract}


\section{INTRODUCTION}

Reconstructing an accurate dense map is crucial for tasks such as autonomous vehicle navigation, robot operation, and virtual reality. Existing dense visual Simultaneous Localization and Mapping (SLAM) frameworks are able to track camera poses and reconstruct complete indoor scenes. However, these methods have always struggled with feature extraction and data association, which cause a serious perceptual aliasing problem \cite{cadena2016past}. Recently, SLAMs leverage Neural Radiance Fields (NeRFs) \cite{mildenhall2021nerf} to operate directly on raw pixel values without designing hand-crafted feature extraction, which omits the above difficulties of traditional methods.

NeRF has recently attracted a lot of research interest, which can obtain more accurate color and precise details by enrolling Multilayer Perceptrons (MLPs). 
Recently, a massive number of works adapt NeRF to SLAM domain, e.g. iMAP \cite{sucar2021imap} applies a neural implicit representation to traditional dense reconstruction. NICE-SLAM \cite{zhu2022nice} adopts a hierarchical and grid-based neural implicit encoding. However, it is required to define the size of reconstruction scenes in advance and to provide a pre-trained CNN model. Vox-Fusion \cite{yang2022vox} exploits octree-based representation and achieves scalable implicit scene reconstruction. The aforementioned methods can reconstruct high-quality maps. However, all these works assume the scenarios are static or deem dynamic objects as outliers. Thus, they are not able to reconstruct the dynamic objects. Naturally, NeRF-based methods are also extended to dynamic scenes marked by the algorithm named D-NeRF \cite{pumarola2021d}, then HyperNeRF \cite{park2021hypernerf} addresses the topological deformation problem in dynamic scenes. However, these works only focus on the mapping side, and directly utilize ground truth camera poses provided by the dataset.

To address these limitations, we propose a novel time-varying implicit representation to track camera poses and reconstruct moving objects in the dynamic scenes, which is named TivNe-SLAM. The inputs of our system are RGB-D image sequences with timestamps. Inspired by D-NeRF \cite{pumarola2021d}, our work extends 3D positions of objects to 4D positions by enrolling time information. Then, we transform points of dynamic objects from the deformation field to the canonical field. Next, colors and Signed Distance Function (SDF) of dynamic scenes are regressed by an MLP. The entire framework simultaneously maintains two processes, including a tracking process and a mapping process, and executes the two processes in turn. To summarize, the contributions of our paper are as follows:
\begin{itemize}
    \item We propose a novel 4D SLAM framework to simultaneously reconstruct the dynamic scenes and estimate the camera poses via Neural Radiance Fields (NeRFs).
    \item We introduce a time-varying representation to capture the position offsets of objects' movement, which enables our framework to eliminate holes and ghost trails that reside in traditional dynamic SLAM frameworks.
    \item We validate a novel overlap-based keyframe selection strategy to reconstruct dynamic objects more completely.
\end{itemize}

The rest of paper is organized as follows: An overview of related work is discussed in Section \ref{sec:rel}. We provide a detailed explanation of our method in Section \ref{sec:method}. In Section \ref{sec:exp}, we demonstrate our experimental results including camera tracking, reconstruction quality, object completion ability, and ablation study. In the end, we summarize the experimental results in Section \ref{sec:conclusion}.

\section{RELATED WORK}\label{sec:rel}
\textbf{Traditional Dense SLAM and Dynamic SLAM:} Classic dense SLAM systems have developed rapidly in the past decades. KinectFusion \cite{izadi2011kinectfusion} introduces a dense reconstruction and tracking system based on Truncated Signed Distance Function (TSDF) with an RGB-D camera. ElasticFusion \cite{whelan2015elasticfusion} is a surface-based SLAM system and proposes two states, active and inactive, to control the activation state of voxels.

However, the above systems solely focus on static scenes, which are impractical in real-world applications. To this end, Co-Fusion \cite{runz2017co} maintains a background model and multiple dynamic foregrounds, detecting moving objects through motion filtering and semantic segmentation. MaskFusion \cite{runz2018maskfusion} utilizes Mask R-CNN \cite{he2017mask} to achieve more precise segmentation of dynamic objects. MID-Fusion \cite{xu2019mid} introduces the Volume TSDF for dense mapping and tracking, but it involves at least four rounds of ray-castings, leading to slow processing. EM-Fusion \cite{strecke2019fusion} proposes a tracking method based on a probabilistic expectation maximization (EM) formulation for reconstructing dynamic objects. TSDF++ \cite{grinvald2021tsdf++} advocates using a single large reconstruction volume to store the entire dynamic scene. RigidFusion \cite{wong2021rigidfusion} leverages motion priors and treats all dynamic objects as one rigid body. ACEFusion \cite{bujanca2022acefusion} adopts a hybrid representation including octrees and surfels. However, all these methods are prone to generating ghost trail effects.

\textbf{NeRF-based Dynamic Scene Reconstruction:}
NeRF is first introduced for static scenes and has recently been extended to dynamic reconstruction. D-NeRF \cite{pumarola2021d} associates time with 3D position and establishes a transformation between the deformation field and the canonical field. Nerfies \cite{park2021nerfies} and HyperNeRF \cite{park2021hypernerf} associate a latent deformation code and an appearance code to each image, realizing a similar transformation to D-NeRF. NSFF \cite{li2021neural} is proposed for handling varieties of in-the-wild scenes, including thin structures, view-dependent effects, and complex degrees of motion. D$^2$NeRF \cite{wu2022d} creatively decouples dynamic objects and a static background for the monocular video. NeRFPlayer \cite{song2023nerfplayer} proposes a decomposition of dynamic scenes based on their temporal characteristics. Park et al. \cite{park2023temporal} utilizes an additional time parameter to execute temporal feature interpolation. PlenOctrees \cite{yu2021plenoctrees} presents a novel octree-based 3D representation to achieve real-time reconstruction, while Fourier-PlenOctrees \cite{wang2022fourier} extends it to dynamic scenes. TiNeuVox \cite{fang2022fast} represents dynamic scenes with optimizable explicit data structures. FFDNeRF \cite{guo2023forward} leverages a forward flow field to better represent object motions. TensoRF \cite{chen2022tensorf} decomposes the 4D scene tensor into multiple low-rank tensors, and HexPlane \cite{cao2023hexplane} decompose 4D volumes in the same way, and represents dynamic scenes as a series of planes. However, all these methods train the models using input images with ground truth camera poses, which impedes their applications in real-world tasks. The method most closely resembling that in our paper is RoDynRF \cite{liu2023robust}, which proposes a space-time synthesis algorithm from a dynamic monocular video and obtains accurate camera poses among high-speed moving objects, but it requires hours of training time.

\textbf{NeRF-based Static SLAM:}
The power of NeRF in synthesizing photo-realistic novel views relies on accurate camera poses. Thus, some researchers integrate NeRF into SLAM. iNeRF \cite{yen2021inerf} is the first work to obtain camera poses by leveraging a carefully trained NeRF. BARF \cite{lin2021barf} further improves iNeRF, by proposing a method to simultaneously train a NeRF and estimate the camera poses. iMAP \cite{sucar2021imap} is the first dense real-time SLAM system based-on NeRFs and is able to estimate camera localization. DROID-SLAM \cite{teed2021droid} is only trained with monocular inputs and directly applied to stereo or RGB-D inputs, obtaining improved accuracy without retraining. NICE-SLAM \cite{zhu2022nice} adopts a local-update strategy and proposes a hierarchical and grid-based neural implicit encoding, but it requires a pre-trained model. Vox-Fusion \cite{yang2022vox} adopts sparse voxel octree, and combines voxel embedding to mitigate artifacts of voxel borders. NICER-SLAM \cite{zhu2024nicer} is the RGB-only dense SLAM system and it proposes a local adaptive transformation for Signed Distance Functions (SDFs). Co-SLAM \cite{wang2023co} designs a joint coordinate and sparse grid encoding. GO-SLAM \cite{zhang2023go} proposes a real-time global pose optimization system that considers the complete history information of input frames and incrementally aligns all poses. Despite the above algorithms achieving decent results for camera localization and scene reconstruction, they all rely on the assumption of a static scene.



\begin{figure*}[ht]
    \centering
    \includegraphics[width=0.99\linewidth]{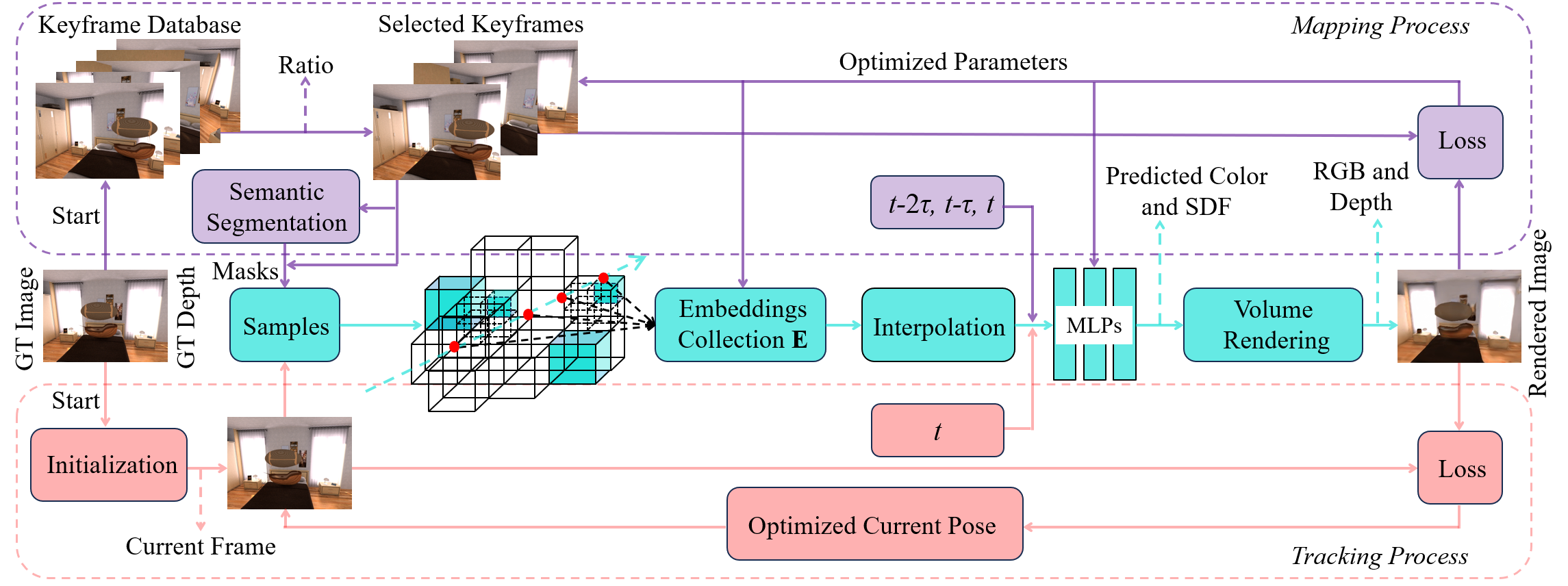}
    \caption{\textbf{Overview of our TivNe-SLAM framework.} Our system simultaneously maintains two processes, a tracking process and a mapping process. \textbf{Tracking Process:} It firstly initializes a map utilizing the 1st frame and initializes the camera pose of each frame. Valid points of the current frame are sampled, encoded by a set of embeddings collection, and tri-linearly interpolated. The interpolated results are fed into two MLPs, and colors and SDF are predicted to render RGB images and depth images. Tracking loss is correspondingly constructed, and mapping parameters are frozen for tracking process, only the pose of the current frame is optimized. \textbf{Mapping Process:} After obtaining the camera pose of the current frame by the tracking process, we design a strategy to select target keyframes from a incrementally-growing database for reconstruction. Then it leverages Mask R-CNN \cite{he2017mask} to obtain the mask segmentation of dynamic objects. As with the tracking process, points sampling, embedding, interpolation, and MLP regression are executed to obtain colors and SDF, and they are used to reconstruct the meshes. The poses, embedding parameters, and MLPs are optimized in the mapping process.}
    \label{fig:overview}
    \vspace{-0.5cm}
\end{figure*}
\section{METHODOLOGY}\label{sec:method}
 Given a stream of continuous RGB-D frames with color images, depths and corresponding timestamps, our TivNe-SLAM simultaneously maintains two processes, a tracking process and a mapping process, the overview is shown in Fig. \ref{fig:overview}. For each frame at time $t$, $N$ pixels are sampled and cast as $N$ rays, and are indexed by $j$. An unbalanced sampling strategy is exploited here, in which we leverage motion masks to distinguish dynamic objects from static background and sample more pixels from dynamic areas. For each ray, $M$ points are further sampled based on density and denoted as $\mathbf{x}_j^i = (x,y,z), i \in \{1,...,M\}, j \in \{1,...,N\}$, where $i$ is the index for sampled 3D points along the $j_{th}$ ray.

\subsection{Deformation Field and Canonical Field}\label{methon:overview}
\begin{figure}[b]
    \centering
    \vspace{-0.5cm}    \includegraphics[width=0.99\linewidth]{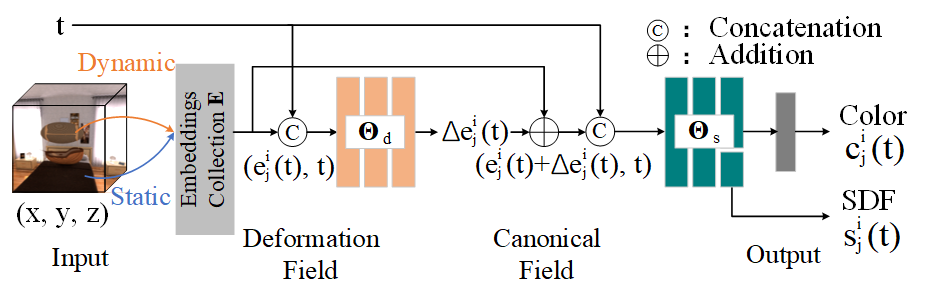}
    \caption{Architecture of our neural deformation field and canonical field. We maintain a sparse embedding collection $\textbf{E} \in \mathbb{R}^{H \times Q}$, which is a collection of Q-Dimensional vectors. We then encode positions $\mathbf{x}^i_j \in \mathbb{R}^3$ as $\mathbf{e}^i_j(t) \in \mathbb{R}^Q$. $\mathbf{e}^i_j(t)$ is associated with time $t$ to regress offsets $\Delta \mathbf{e}^i_j(t)$ by $\Theta_{d}$. Colors and SDFs are obtained by feeding $(\mathbf{e}^i_j(t) + \Delta \mathbf{e}^i_j(t), t)$ to $\Theta_{s}$.}
    \label{fig:MLP}
\end{figure}
The detailed architecture of our neural fields transformation is further shown in Fig. \ref{fig:MLP}. We define the scene at $t = 0$ as the canonical field and the scene at the other time as the deformation field. Inspired by D-NeRF \cite{pumarola2021d}, we leverage a deformation neural network $\Theta_{d}$ to transform positions from the deformation field to the canonical field. Specifically, $\Theta_{d}$ is trained to regress the offsets which represent moving distance of every point between the deformation field and the canonical field. To be noticed, the offset is zero only if $t = 0$. $\mathbf{e}^i_j(t)$ is tri-linearly interpolated result from embeddings collection \textbf{E} and concatenated as $(\mathbf{e}^i_j(t), t)$ by adding a timestamp. The function of the deformation field can be expressed as Equation (\ref{eq:deltad}):
\begin{align}
\Theta_{d}(\mathbf{e}^i_j(t),t)=\begin{cases}
 \Delta \mathbf{e}^i_j(t) & \text{ if } t\neq0, \\
 0 & \text{ if } t=0.
\end{cases}
\label{eq:deltad}
\end{align}

Accordingly, we employ another MLP, $\Theta_{s}$, to regress colors and SDFs. Considering SDF values,  instead of volume densities, are required in our task, the volume rendering formula utilized by the original NeRF \cite{mildenhall2021nerf} needs to be improved. Thus, we enroll a volume rendering technique via combining the network architecture in Azinovi{\'c} et al. \cite{azinovic2022neural} and Vox-Fusion \cite{yang2022vox}. 

In order to represent dynamic scenes, we further modify the formulas as follows:
\begin{gather}\label{eq:coloransdf}
   \mathbf{e}^i_j(t)=\text{TriLerp}(\hat{\xi}(t)T(t-\tau)\mathbf{u}(t), \mathbf{E}).\\
    \mathbf{c}^i_j(t),s^i_j(t)=\Theta_s(\Theta_{d}({e}^i_j(t),t)+\mathbf{e}^i_j(t),t).
\end{gather}
where $\text{TriLerp}(\cdot, \cdot)$ denotes the tri-linear interpolation function which yields interpolated embeddings $\mathbf{e}^i_j(t)$. $\mathbf{u}(t)=\{(u,v)\}$ denotes pixels of current frame $t$. $T(t-\tau) \in SE(3)$ is the pose of the previous frame at time $t - \tau$, and $\hat{\xi}(t) \in SE(3)$ denotes the relative pose of the current frame and the previous frame. By following a zero-motion model, we always initialize the pose of the current frame with the pose of the last frame.
Additionally, $\mathbf{c}^i_j(t)$ and $s^i_j(t)$ are the colors and SDFs of these points along each camera ray respectively.
\begin{gather}\label{eq:weight}
    \omega^i_j(t) = \sigma(\frac{s^i_j(t)}{tr})\cdot \sigma(-\frac{s^i_j(t)}{tr}).
\end{gather}
Where ${tr}$ is a predefined parameter which presents truncated signed distance, $\sigma(\cdot)$ is the sigmoid activation function. Weight $\omega^i_j(t)$ gets the maximum value when $\sigma(\frac{s^i_j(t)}{tr})=\sigma(-\frac{s^i_j(t)}{tr})$, which means the point is around the object surface.

Lastly, $\mathbf{c}^i_j(t)$ and ${d^i_j(t)}$ stand for the colors and z-axis value of the $i_{th}$ point along the $j_{th}$ ray. The rendered color and rendered depth are calculated through the weighted summation, as shown in $\mathbf{C}_j(t)$ and $D_j(t)$:
\begin{gather}\label{eq:render}
    \mathbf{C}_j(t) = \frac{1}{\sum^{M-1}_{i=0}\omega^i_j(t)}\sum^{M-1}_{i=0}\omega^i_j(t)\cdot\mathbf{c}^i_j(t).\\
    D_j(t) = \frac{1}{\sum^{M-1}_{i=0}\omega^i_j(t)}\sum^{M-1}_{i=0}\omega^i_j(t)\cdot d^i_j(t).
\end{gather}

\subsection{Loss Function}
Within each frame, N pixels are sampled and are further divided into $B$ batches. Each batch is denoted as $b$. Each pixel is corresponding to a ray-casting. We redesign the loss function used in Azinovi{\'c} et al \cite{azinovic2022neural} and Vox-Fusion \cite{yang2022vox} as following: 
\begin{gather}\label{eq:loss}
\begin{aligned}
    \mathcal{L}^b(t) = &\sum_{b=0}^{B-1}(\lambda_{color}\mathcal{L}^b_{color}(t) 
     + \lambda_{depth}\mathcal{L}^b_{depth}(t) \\
     &+ \lambda_{space}\mathcal{L}^b_{space}(t) 
     + \lambda_{SDF}\mathcal{L}^b_{SDF}(t)\\
     &+ \lambda_{offset}\mathcal{L}^b_{offset}(t)).
\end{aligned}
\end{gather}
where $\mathcal{L}^b(t)$ is made up by five terms: color loss, depth loss, free-space loss, SDF loss, and offset loss. Finally, they are multiplied by their individual coefficients and are added together. 

Specifically, color loss $\mathcal{L}^b_{color}(t)$ and depth loss $\mathcal{L}^b_{depth}(t)$ of sampling points are defined as:
\begin{gather}\label{eq:losscoloranddepth}
    \mathcal{L}^b_{color}(t) = \frac{1}{N}\sum_{j=0}^{N-1}\lVert \textbf{C}_j(t) - \textbf{C}^{gt}_j(t)\rVert.\\
    \mathcal{L}^b_{depth}(t) = \frac{1}{N}\sum_{j=0}^{N-1}\lVert D_j(t) - D^{gt}_j(t)\rVert.
\end{gather}
We predict color and depth at time t using $\Theta_s$, and the loss between predicted images (RGB and depth) and the ground-truth images is constructed.

Then, free-space loss $\mathcal{L}^b_{space}(t)$ and SDF loss $\mathcal{L}^b_{SDF}(t)$ are defined as:
\begin{equation}
\begin{split}
    \mathcal{L}^b_{space}(t) = \beta_{space} \sum_{s \in S_{space}}(D_s(t) - tr)^2,\\
    where \quad \beta_{space} = (1 - \frac{P_{space}(t)}{P_{space}(t)+P_{tr}(t)}).
\end{split}
\end{equation}
\begin{equation}
\begin{split}
    \mathcal{L}^b_{SDF}(t) = \beta_{SDF} \sum_{s \in S_{tr}}(D_s(t) - {D}_s^{gt}(t))^2,\\
    where \quad \beta_{SDF} = (1 - \frac{P_{tr}(t)}{P_{space}(t)+P_{tr}(t)}).
\end{split}
\end{equation}
$P_{space}(t)$ and $P_{tr}(t)$ are the number of points in free space and near the surface of the object respectively. $s \in S_{space}$ indicate those points sampled in the free space (i.e. in front of the object surface) and $s \in S_{tr}$ are the points near to object surface. In $\mathcal{L}^b_{space}(t)$, $D_s(t)$, the depth of $s \in S_{space}$, is trained to be equal to truncated distance $tr$. In $\mathcal{L}^b_{SDF}(t)$, $D_s(t)$, the depth of $s \in S_{tr}$, is restricted within the truncation range and learns to be equal to ground truth value ${D}_s^{gt}(t)$.

Since the position of the background is supposed to remain static, the $\mathcal{L}^b_{offset}(t)$ is designed to minimize the offset of the background, as shown below: 
\begin{equation}
\begin{split}
    \mathcal{L}^b_{offset}(t) = \sum_{s \in S_{bg}}\lVert \Delta e(t)\rVert ^2.
\end{split}
\end{equation}
$s \in S_{bg}$ are the sampled points from the background. $\Delta e(t)$ is the offset embeddings defined in Equation (\ref{eq:deltad}). 

\subsection{Camera Tracking}
\subsubsection{\textbf{Generation of Motion Mask}}
In dynamic scenes, moving objects disturb the estimation of camera poses. Ours is able to estimate a precise camera pose and reconstruct the dynamic scenes by leveraging masks of moving objects. Our design is robust enough that it can work with a vanilla Instance Segmentation (IS) module as simple as Mask R-CNN \cite{he2017mask}.

\subsubsection{\textbf{Estimation of Camera Pose}}
In the tracking process, we freeze the parameters of embeddings collection $\mathbf{E}$ and two MLPs ($\Theta_{d}$ and $\Theta_{s}$), and only optimize the pose $T(t) = \hat{\xi}(t)T(t-\tau)$ of current frame. $\hat{\xi}(t) \in SE(3)$ is the relative pose from the last frame to the current frame which is defined in Equation (\ref{eq:coloransdf}). Specifically, we sample rays in the whole input image, then obtain the predicted colors and SDFs. In the end, we compute the loss described in Equation (\ref{eq:loss}) between the predicted images and the ground-truth images. 
\begin{figure*}[t]
  \centering
  \setlength{\tabcolsep}{1pt}
  \begin{tabular}{ccccc}
    &\multicolumn{1}{c}{ NICE-SLAM \cite{zhu2022nice}} & \multicolumn{1}{c}{ Vox-Fusion \cite{yang2022vox}} & \multicolumn{1}{c}{ Ours (Random)} & \multicolumn{1}{c}{ Ours (Overlap)}\\
    \makecell{\rotatebox{90}{Room4-1}}&
    \makecell{\includegraphics[width=.24\linewidth]{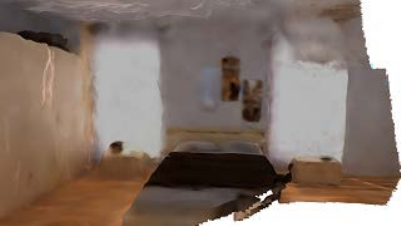}}&
    \makecell{\includegraphics[width=.24\linewidth]{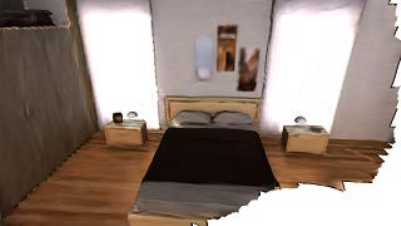}}&
    \makecell{\includegraphics[width=.24\linewidth]{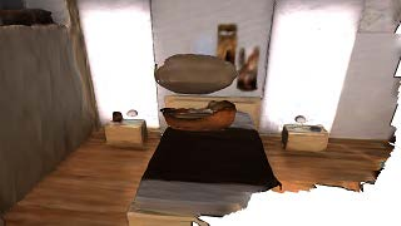}}&
    \makecell{\includegraphics[width=.24\linewidth]{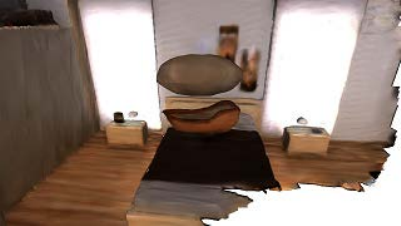}}\hspace{2pt}\vspace{-2pt} \\
    \makecell{\rotatebox{90}{Room4-2}}&
    \makecell{\includegraphics[width=.24\linewidth]{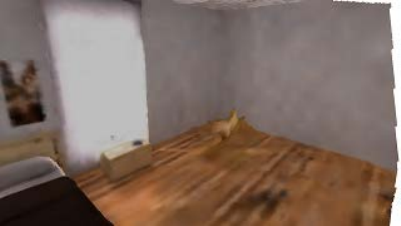}}&
    \makecell{\includegraphics[width=.24\linewidth]{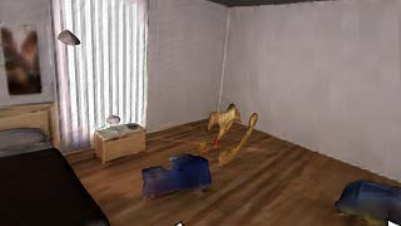}}&
    \makecell{\includegraphics[width=.24\linewidth]{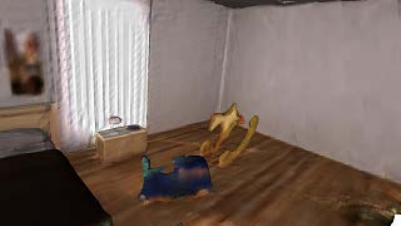}}&
    \makecell{\includegraphics[width=.24\linewidth]{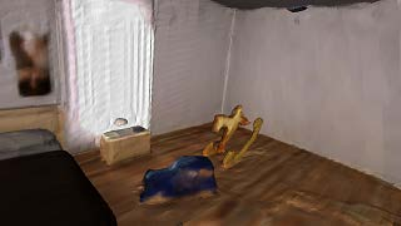}}\hspace{2pt}\vspace{-2pt}\\
    \makecell{\rotatebox{90}{ToyCar3}}&
    \makecell{\includegraphics[width=.24\linewidth]{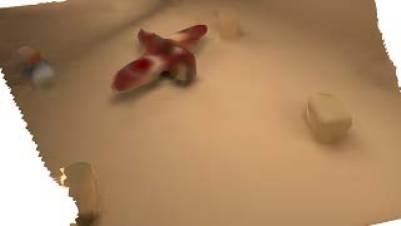}}&
    \makecell{\includegraphics[width=.24\linewidth]{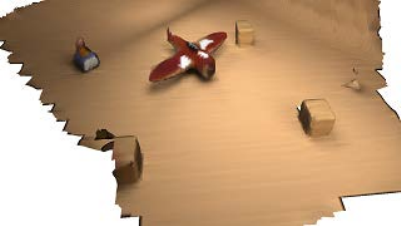}}&
    \makecell{\includegraphics[width=.24\linewidth]{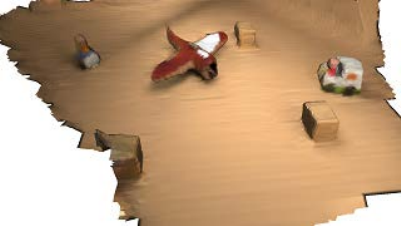}}&
    \makecell{\includegraphics[width=.24\linewidth]{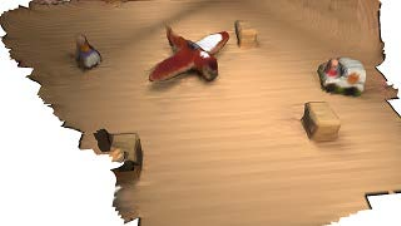}}\hspace{2pt}\vspace{-2pt}\\
    \makecell{\rotatebox{90}{Teddy}}&
    \makecell{\includegraphics[width=.24\linewidth]{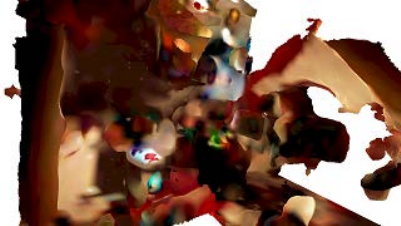}}&
    \makecell{\includegraphics[width=.24\linewidth]{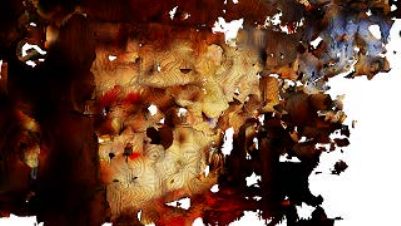}}&
    \makecell{\includegraphics[width=.24\linewidth]{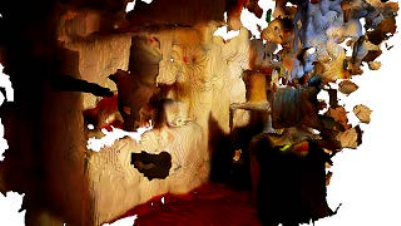}}&
    \makecell{\includegraphics[width=.24\linewidth]{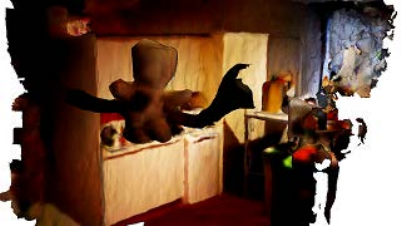}}\hspace{2pt}
  \end{tabular}
  \caption{\textbf{Comparison of Mesh Quality among Two NeRF-based SLAMs and Two keyframe-selection Variations of Our TivNe-SLAM.} Mapping results of different datasets are interpreted row by row. \textbf{(1) Room4-1: }We completely reconstruct the flying ship. However, NICE-SLAM \cite{zhu2022nice} and Vox-Fusion \cite{yang2022vox} are unable to capture it. Randomly selecting keyframes generates unstable shapes and brings gaps into the reconstructed objects. Overlap-based TivNe-SLAM handles this problem well. \textbf{(2) Room4-2:} The results indicate that NICE-SLAM cannot reconstruct the blue car at all, and Vox-Fusion can only occasionally reconstruct the dynamic car, and fails to eliminate residual reconstruction in history positions. Similarly, our method with overlap-based strategy generates the best results. \textbf{(3) ToyCar3:} Only our method reconstructs the white car on the right side of the scene, but the others treat it as an outlier. Additionally, the reflected light on the plane's wings is gradually shifted to the left wing in the input image, but our method still effectively captures this transitional process. \textbf{(4) Teddy:} It is clear that neither NICE-SLAM nor Vox-Fusion can reconstruct the scene of this real-world dynamic dataset. The method of exploiting randomly-selected keyframes yields poor results. However, our method based on overlapping selection fully reconstructs the dynamic teddy, arms of a person and the static background. (The brightness of images of the Teddy dataset is slightly adjusted for clearer visualization.) }
    \label{fig:compare}
    \vspace{-0.7cm}
\end{figure*}

\subsection{Dynamic Mapping}
\subsubsection{\textbf{Keyframe Selection}}
We maintain a keyframe database and insert new keyframes at a fixed interval. Vox-Fusion \cite{yang2022vox} randomly selects keyframes from the keyframe database to optimize the global map, which leads to incomplete reconstruction.
We propose a novel strategy to select keyframes from the database by calculating the overlapping ratio between the current frame and each frame in the database. Specifically, we reproject the pixel points, $\mathbf{u}(t)$, sampled from the current frame $t$ back into 3D space, and project them into the 2D plane of the target keyframe, as shown in Equation (\ref{eq:overlap}). Then we calculate whether $\mathbf{u}(t- q \tau)$ falls inside the pixel coordinate of the  keyframe at $t - q \tau$, where $q$ denotes the number of interval $\tau$. We determine the overlapping ratio by computing the proportion of valid points to the total sampled points. Keyframes with a lower overlapping ratio are picked up for mapping optimization, causing the selected keyframes to cover as many novel views of scenes as possible. 
\begin{gather}\label{eq:overlap}
    \mathbf{u}(t-q \tau)= K \cdot T_{w2c}(t - q \tau) \cdot T_{c2w}(t) \cdot K^{-1} \cdot\mathbf{u}(t).
\end{gather}
Where $K$ represents the intrinsic matrix of the camera. $T_{c2w}$ and $T_{w2c}$ are the transformation matrix between the camera coordinate and the world coordinate.

\subsubsection{\textbf{Optimization \& Visualization}}
Unlike the tracking process, the mapping process involves the overall optimization of parameters of embedding collections $\mathbf{E}$, MLPs ($\Theta_{d}$ and $\Theta_{s}$), and the poses of target keyframes and the current frame. The loss function adopted for optimization is identical to Equation (\ref{eq:loss}). After self-supervised training of each frame, we obtain an octree-based map up to the current frame, the newly trained MLP can regress the colors and SDF of every voxel in the scene. Then the mesh can be generated in real time using the marching cubes \cite{lorensen1998marching} algorithm.



\section{Experiments}\label{sec:exp}
\subsection{Experimental Setup}
\subsubsection{\textbf{Implementation Details}}
We implement two tiny MLPs, $\Theta_d$ and $\Theta_s$, in our experiment. Both of the MLPs consist of three hidden layers, and one Fully-Connected (FC) layer is attached for output of $\Theta_s$. The SDFs are obtained from the third hidden layer of $\Theta_s$ and the color is obtained from the FC layer of $\Theta_s$. The matrix size of embedding collection \textbf{E} is configured as $20000 \times 16$. Moreover, the tracking process of a single frame is configured to run for 30 iterations and the mapping process for 20 iterations. The camera pose of the first frame is initialized as an identity matrix. The size of input images for training are $640 \times 480$. We down-sample the image to $320 \times 240$ for visualization of our rendering experiments. All experiments' results are evaluated on an NVIDIA RTX 4090 GPU card with 24 GB memory.

\subsubsection{\textbf{Datasets}}
We evaluate our system on two different public synthetic datasets (i.e. Room4 and ToyCar3) and a real-world dataset (i.e. Teddy) which are provided by Co-Fusion \cite{runz2017co}. All three Datasets saved as RGB-D sequences with dynamic objects. Additionally, these datasets provide corresponding videos captured by a Kinect camera, from which we extract timestamps.
\subsection{Evaluation of Camera Tracking}

Absolute Trajectory Error (ATE) is adopted to evaluate the accuracy of camera tracking. We compare our system with other NeRF-based SLAM systems (i.e. NICE-SLAM and Vox-Fusion), dynamic NeRF (i.e. RoDynRF) and Structure-from-Motion technique (i.e. COLMAP \cite{schonberger2016structure}). As shown in TABLE \ref{table:ATE}, best results are highlighted
as \colorbox{light}{\bf first} and \colorbox{lightlight}{second}. We can clearly observe that our system predicts the most accurate camera pose in dynamic scenes compared to other methods on the Room4 dataset and the ToyCar3 dataset. That is to say, we successfully reconstruct the dynamic objects in the scenes, which is also beneficial for estimating more precise camera poses. Unfortunately, the real-world dataset lacks of ground truth camera trajectory, so that ATE of Teddy dataset is not evaluated.
\begin{table}[ht]
\vspace{0.2cm}
\caption{\small Camera Tracking Results}
\label{table:ATE}
\begin{center}
\begin{tabular}{c||c||>{\centering\arraybackslash}p{1.5cm}|>{\centering\arraybackslash}p{1.5cm}}
\hline
Methods&ATE& Room4& ToyCar3\\
\hline
\multirow{3}{*}{\makecell{COLMAP \cite{schonberger2016structure}}}
&RMSE [m]($\downarrow$) & 0.3448 & 0.1187\\ 
&Mean [m]($\downarrow$) & 0.2882 & 0.1039\\
&Std [m]($\downarrow$) & 0.1892 & 0.0574\\
\hline
\multirow{3}{*}{\makecell{NICE-SLAM \cite{zhu2022nice}}} 
&RMSE [m]($\downarrow$) & 0.1305 & 0.0785\\ 
&Mean [m]($\downarrow$) & 0.0680 & 0.0617\\
&Std [m]($\downarrow$) & 0.1113 & 0.0486\\
\hline
\multirow{3}{*}{\makecell{Vox-Fusion \cite{yang2022vox}}}
&RMSE [m]($\downarrow$) &\second 0.0164 &\second0.0655\\ 
&Mean [m]($\downarrow$) & \second0.0147 & \second0.0543\\
&Std [m]($\downarrow$) & \second0.0073 & \second0.0366\\
\hline
\multirow{3}{*}{\makecell{RoDynRF \cite{liu2023robust}}}
&RMSE [m]($\downarrow$) & 0.3903 & 0.2338\\ 
&Mean [m]($\downarrow$) & 0.3408 & 0.2051\\
&Std [m]($\downarrow$) & 0.1903 & 0.1123\\
\hline



\multirow{3}{*}{\makecell{Ours}}
&RMSE [m]($\downarrow$) & \fst 0.0131 & \fst0.0559\\ 
&Mean [m]($\downarrow$) & \fst0.0118 & \fst0.0444\\
&Std [m]($\downarrow$) & \fst0.0056 & \fst0.0339\\

\hline
\end{tabular}
\end{center}
\vspace{-0.5cm}
\end{table}

\subsection{Evaluation of Mapping Quality}
\subsubsection{\textbf{Qualitative Analysis}} \label{subsec:rt}
Qualitative comparison of the reconstruction results between NICE-SLAM, Vox-Fusion, RoDynRF and our system (TivNe-SLAM) is shown in Fig. \ref{fig:compare}. To be noticed, RoDynRF, Vox-Fusion and our TivNe-SLAM do not require pre-trained geometric models, while NICE-SLAM does. From the experimental results of all three datasets, obviously, we achieve better reconstruction quality for dynamic objects. Only our method successfully reconstructs all dynamic objects, including the flying ship, the rocking horse and the moving blue car of the Room4 dataset, the moving white car of the ToyCar3 dataset, and the moving teddy bear of the Teddy dataset. Especially in the Teddy dataset, other methods fail to reconstruct the scene when dynamic objects appear. NICE-SLAM is able to reconstruct scenes, but it still treats the dynamic objects as outliers and it tries to avoid reconstructing these dynamic objects as much as possible. Occasionally, Vox-Fusion only partially and inaccurately reconstructs dynamic objects and there are also some dynamic objects that are not reconstructed at all. We will further analyze the differences between the two different keyframe selection strategies in Section. \ref{subsec:complete}.

\begin{figure}[h]
  \centering
  \setlength{\tabcolsep}{0.7pt}
  \begin{tabular}{cccc}
    &\multicolumn{1}{c}{Ground Truth} & \multicolumn{1}{c}{ RoDynRF \cite{liu2023robust}} & \multicolumn{1}{c}{ Ours} \vspace{-0.1em} \\
     \makecell{\rotatebox{90}{Room4}}  &
     \makecell{\includegraphics[width=.31\linewidth]{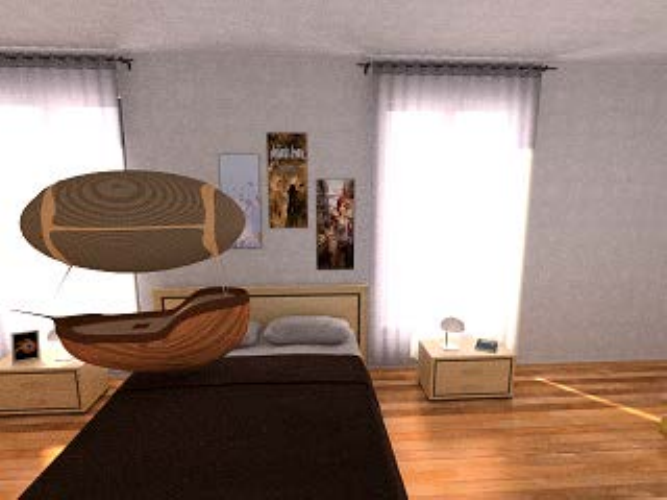}} & 
    \makecell{\includegraphics[width=.31\linewidth]{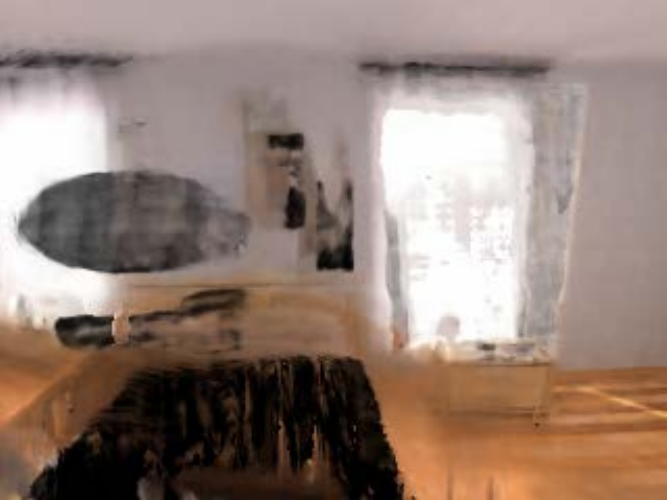}} &
    \makecell{\includegraphics[width=.31\linewidth]{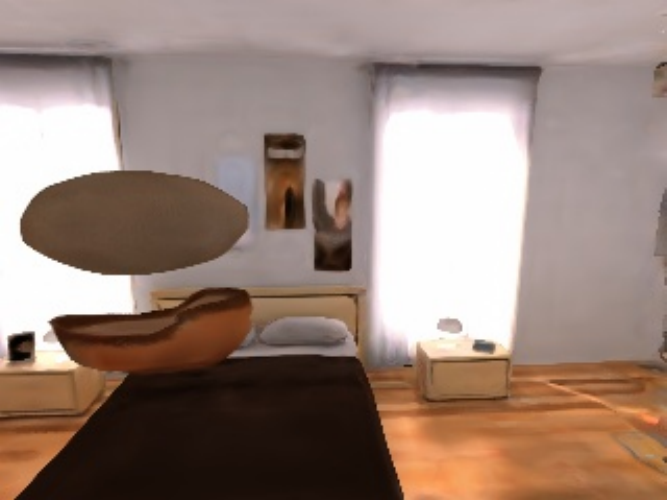}} \vspace{-2pt}\\
    \makecell{\rotatebox{90}{ToyCar3}}  &
    \makecell{\includegraphics[width=.31\linewidth]{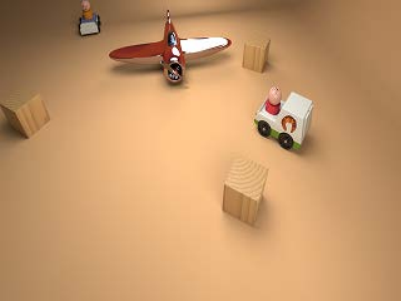}} & 
    \makecell{\includegraphics[width=.31\linewidth]{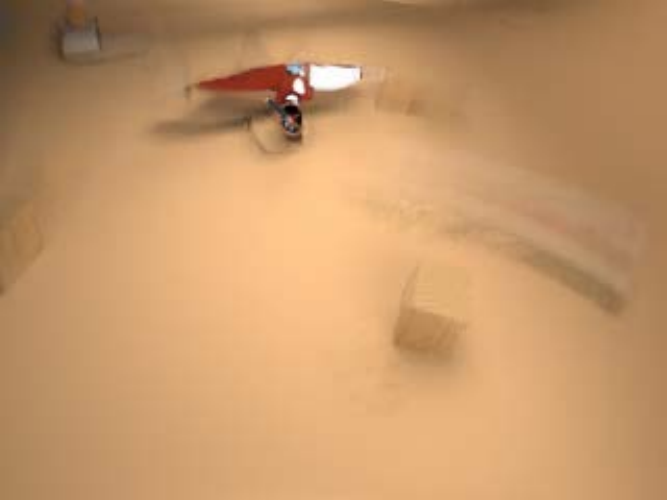}}&
    \makecell{\includegraphics[width=.31\linewidth]{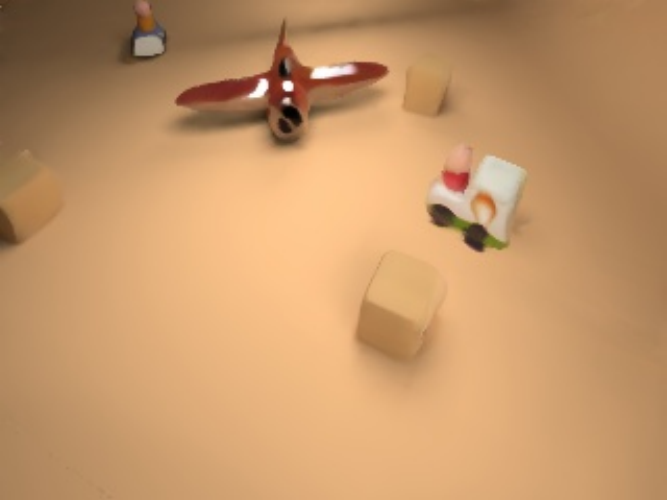}} \vspace{-2pt}\\
    \makecell{\rotatebox{90}{Teddy}}  &
    \makecell{\includegraphics[width=.31\linewidth]{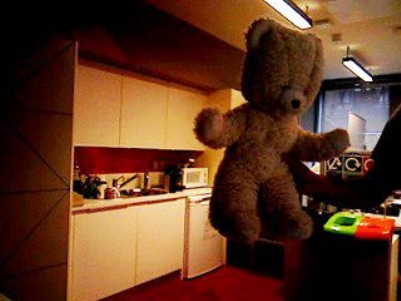}} & 
    \makecell{\includegraphics[width=.31\linewidth]{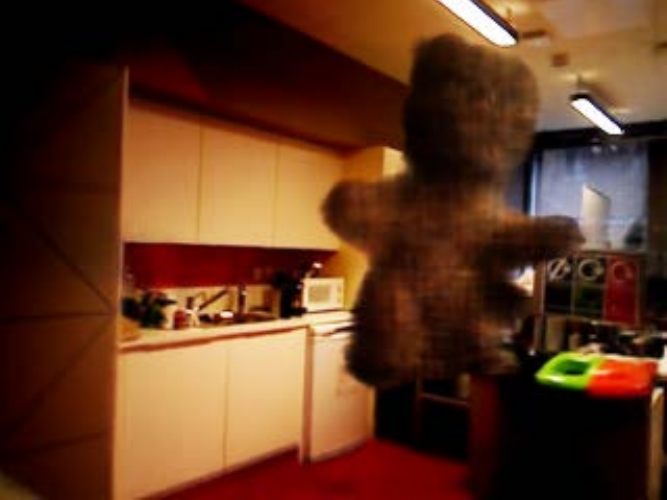}}&
    \makecell{\includegraphics[width=.31\linewidth]{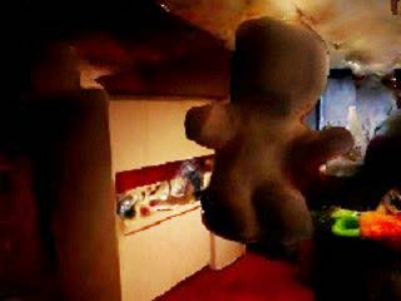}} 
    \vspace{-2pt}
  \end{tabular}
    \caption{\textbf{Results Comparison between Rendered Images of Dynamic NeRF.} Our TivNe-SLAM precisely renders images that closely resemble the input images on all three scenes. However RoDynRF fails to completely render dynamic objects in the first two scenes with fast-moving views. For the third scene, in which the camera moves more slowly, RoDynRF renders better result for both dynamic objects and the static background by adding the training time.}
    \label{fig:dynerf}
    \vspace{-0.3cm}
\end{figure}
As shown in Fig. \ref{fig:dynerf}, we compare the images rendered by RoDynRF and our TivNe-SLAM. Our system achieves superior rendering results on synthetic datasets, because the camera moves rapidly in these scenarios, the views are switched at a fast speed. RoDynRF fails to accurately estimate camera poses and geometry because it is designed for scenes with slowly moving camera views.
For the Teddy dataset, which is a real-world dataset, our system achieves competitive rendering results. Our method is able to reconstruct dynamic objects well, but it has difficulty rendering static backgrounds as well as the RoDynRF method. However, our method has an advantage in training time, the RoDynRF uses around 40 to 50 times as many training hours as ours. This is further reported and analyzed in Section \ref{subsec:MapQttA}.

\begin{figure}[ht!]
  \centering
  \setlength{\tabcolsep}{0.7pt}
  \begin{tabular}{ccc}
    \multicolumn{1}{c}{ Co-Fusion \cite{runz2017co}} & \multicolumn{1}{c}{ MaskFusion \cite{runz2018maskfusion}} & \multicolumn{1}{c}{ Ours}  \\
     \makecell{\includegraphics[width=.32\linewidth]{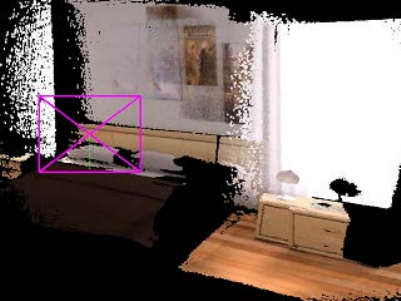}} & 
    \makecell{\includegraphics[width=.32\linewidth]{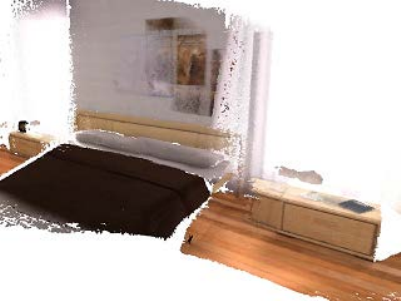}} &
    \makecell{\includegraphics[width=.32\linewidth]{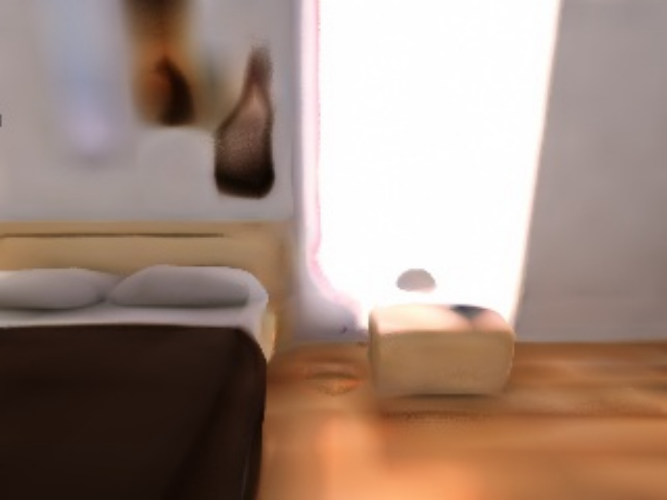}} \vspace{-2pt}\\
    
    \makecell{\includegraphics[width=.32\linewidth]{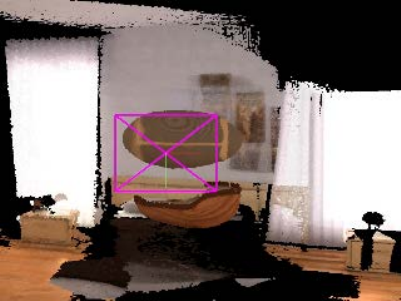}} & 
    \makecell{\includegraphics[width=.32\linewidth]{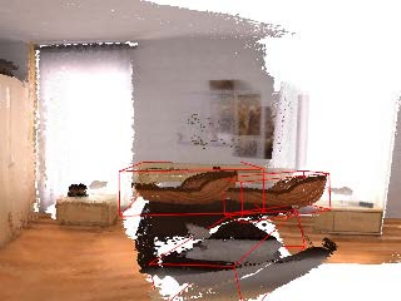}}&
    \makecell{\includegraphics[width=.32\linewidth]{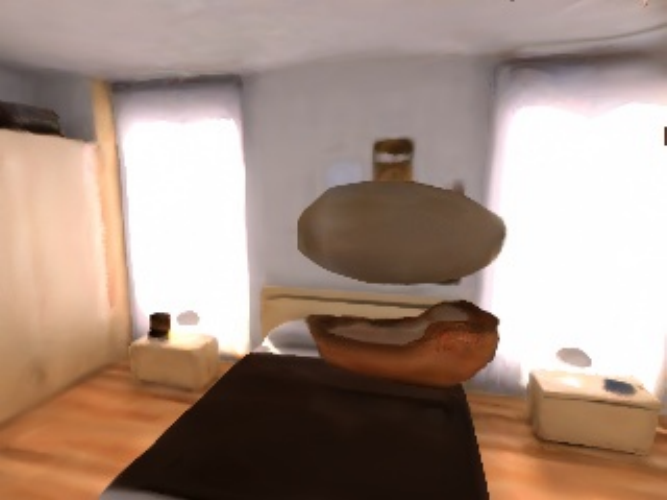}} 
    
    \vspace{-2pt}
  \end{tabular}
    \caption{\textbf{Results Comparison with Traditional Dynamic SLAM Systems.} We also compare our method with the Co-Fusion and MaskFusion, which both rely on a pre-trained CNN model. Both methods produce more artifacts (e.g., black bed and flying ship) and gaps (black and white regions) in the reconstructed scenes compared to our framework.}
    \label{fig:tradition}
    \vspace{-0.3cm}
\end{figure}

Lastly, we also compare our mapping results with traditional SLAM methods which are capable of pose estimation. We demonstrate that these methods bring a series of reconstruction problems like ghost trails and empty holes for dynamic scenes. Results is shown in Fig. \ref{fig:tradition}. This verifies the superiority of our method to enroll dynamic NeRF to a traditional SLAM.
\subsubsection{\textbf{Quantitative Analysis}}
\label{subsec:MapQttA}
 Quantitative results with four metrics, \textbf{MSE} (Mean Squared Error), \textbf{PSNR} (Peak Signal-to-Noise Ratio), \textbf{SSIM}  (Structural Similarity Index) and \textbf{LPIPS}  (Learned Perceptual Image Patch Similarity) \cite{zhang2018unreasonable} are shown in TABLE \ref{table:nerf} with best results highlighted
as \colorbox{light}{\bf first} and \colorbox{lightlight}{second}. It should be noted in advance that the scores of Vox-Fusion on the Teddy dataset are not reported, because scene reconstruction of Vox-Fusion crashes when dynamic objects appear in the scene.


As seen in TABLE \ref{table:nerf}, our system achieves the best rendering results in comparison to the other three methods on two synthetic datasets. On the Teddy dataset, ours also obtains competitive results. The reason we cannot achieve the best results on real datasets is that we only used two light-weight MLPs to train the entire scene. When the scene is relatively large or complex, two tiny MLPs cannot completely regress the entire scene. 
\begin{table}[ht]
\vspace{0.2cm}

\caption{\small Novel View Synthesis Evaluation}
\label{table:nerf}
\begin{center}
\begin{tabular}{c||c||>{\centering\arraybackslash}p{1cm}|>{\centering\arraybackslash}p{1cm}|>{\centering\arraybackslash}p{1cm}}
\hline
Methods& Metrics & Room4& ToyCar3& Teddy\\
\hline
\multirow{4}{*}{\makecell{NICE-SLAM \cite{zhu2022nice}}} 
& MSE($\downarrow$) & 0.0114 & 0.0045 &0.0148\\ 
&PSNR($\uparrow$) & 22.1614 & 24.1507&18.4256\\
&SSIM($\uparrow$) & 0.6755 & 0.8950& 0.5483\\
&LPIPS($\downarrow$) & 0.5983& 0.2851&0.5420\\

\hline
\multirow{4}{*}{\makecell{Vox-Fusion \cite{yang2022vox}}}
& MSE($\downarrow$) & \second0.0040 & \second0.0033&N/A\\ 
&PSNR($\uparrow$) & \second25.5220 & \second25.1758&N/A\\
&SSIM($\uparrow$) & \second0.7882 & \second0.9158&N/A\\
&PLIPS($\downarrow$) & \second0.3969 & \second0.1891&N/A\\
\hline
\multirow{4}{*}{\makecell{RoDynRF \cite{liu2023robust}}}
& MSE($\downarrow$) & 0.0095 & 0.0047&\fst0.0012\\ 
&PSNR($\uparrow$) & 21.0041 & 23.2258&\fst29.5681\\
&SSIM($\uparrow$) & 0.6146 & 0.8875&\fst0.8673\\
&PLIPS($\downarrow$) & 0.5352 & 0.2727&\fst0.1764\\
\hline
\multirow{4}{*}{\makecell{Ours (overlap)}}
& MSE($\downarrow$) & \fst 0.0020 & \fst0.0032&\second0.0069\\ 
&PSNR($\uparrow$) & \fst27.2393 & \fst25.3456& \second22.6057\\
&SSIM($\uparrow$) & \fst0.8103 & \fst0.9213 &\second0.7099\\
&LPIPS($\downarrow$) & \fst0.3711 & \fst0.1604&\second0.3802\\
\hline
\end{tabular}
\end{center}
\vspace{-0.5cm}

\end{table} 

However, as mentioned in Section. \ref{subsec:rt}, the decent rendering results of RoDynRF requires extensively time-consuming self-supervised training which prolongs the map building time. We report the approximated time required to train RoDynRF and TivNe-SLAM on all three datasets in TABLE \ref{table:time}, and it is evident that our method trains much faster than RoDynRF on all datasets. Compared to RoDynRF, our system not only ensures a certain level of capture of dynamic scenes but also achieves real-time performance.

\begin{table}[ht]
\caption{Comparison of Training \& Mapping Time for Dynamic NeRF}
\label{table:time}
\begin{center}
\begin{tabular}{c||c||>{\centering\arraybackslash}p{4.0cm}}
\hline
Methods&Datasets&Training \& Mapping Time ($\downarrow$)\\
\hline
\multirow{3}{*}{\makecell{RoDynRF \cite{liu2023robust}}} 
&Room4& \textgreater 11h\\ 
&ToyCar3& \textgreater 10h\\
&Teddy& \textgreater 12h\\
\hline
\multirow{3}{*}{\makecell{Ours}} 
&Room4& \textless 0.3h\\ 
&ToyCar3& \textless 0.2h\\
&Teddy& \textless 0.3h\\
\hline
\end{tabular}
\end{center}
\vspace{-0.8cm}
\end{table}

\subsection{Object Completion}
\label{subsec:complete}
The keyframe selection strategy used in Vox-Fusion \cite{yang2022vox} is unstable, so its experimental results are different across multiple executions of experiments. 
Thus, we propose a novel keyframe selection strategy, calculating the overlap between current frame and the keyframes database. Then keyframes with a lower overlapping ratio are selected to reconstruct the dynamic scene. As expected, we achieve outstanding completeness in reconstructing dynamic objects.
\begin{figure}[h!]
  \centering
  \setlength{\tabcolsep}{10pt}
  \begin{tabular}{cc}
    \multicolumn{1}{c}{Random Selection} & \multicolumn{1}{c}{ Overlap-Based Selection} \\
    \makecell{\includegraphics[width=.3\linewidth]{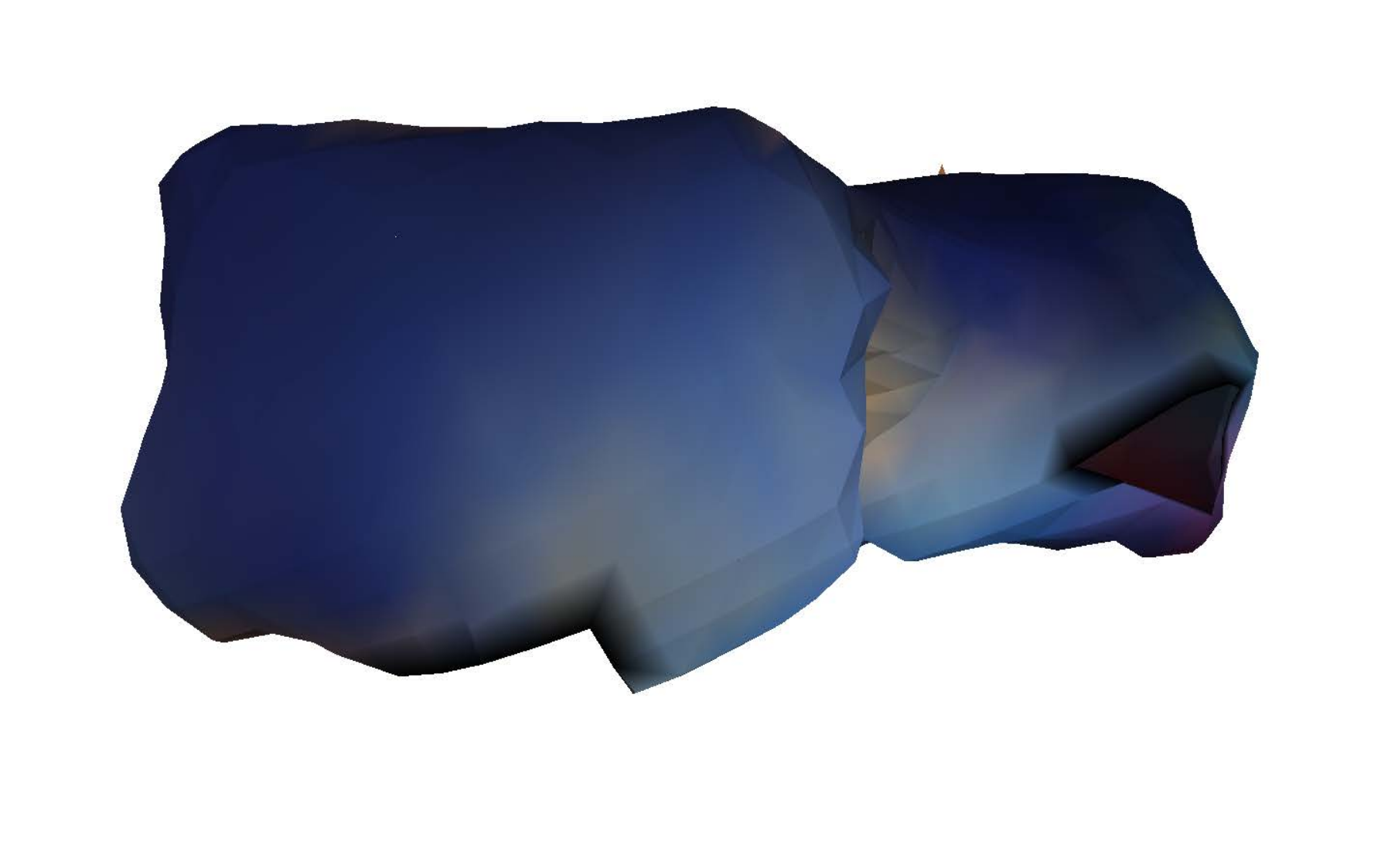}} & 
    \makecell{\includegraphics[width=.3\linewidth]{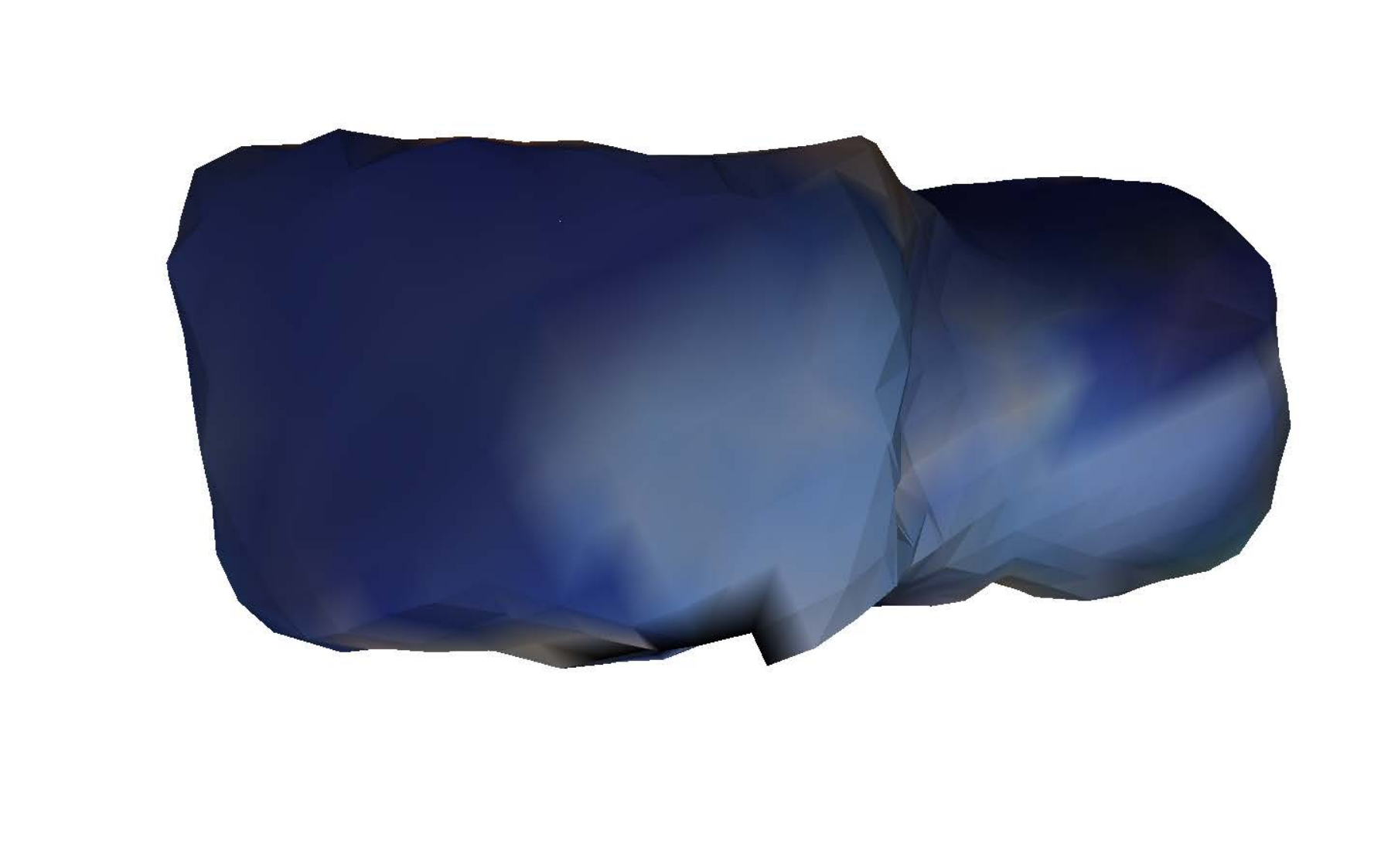}} \vspace{-0.3em}\\
    \makecell{\includegraphics[width=.3\linewidth]{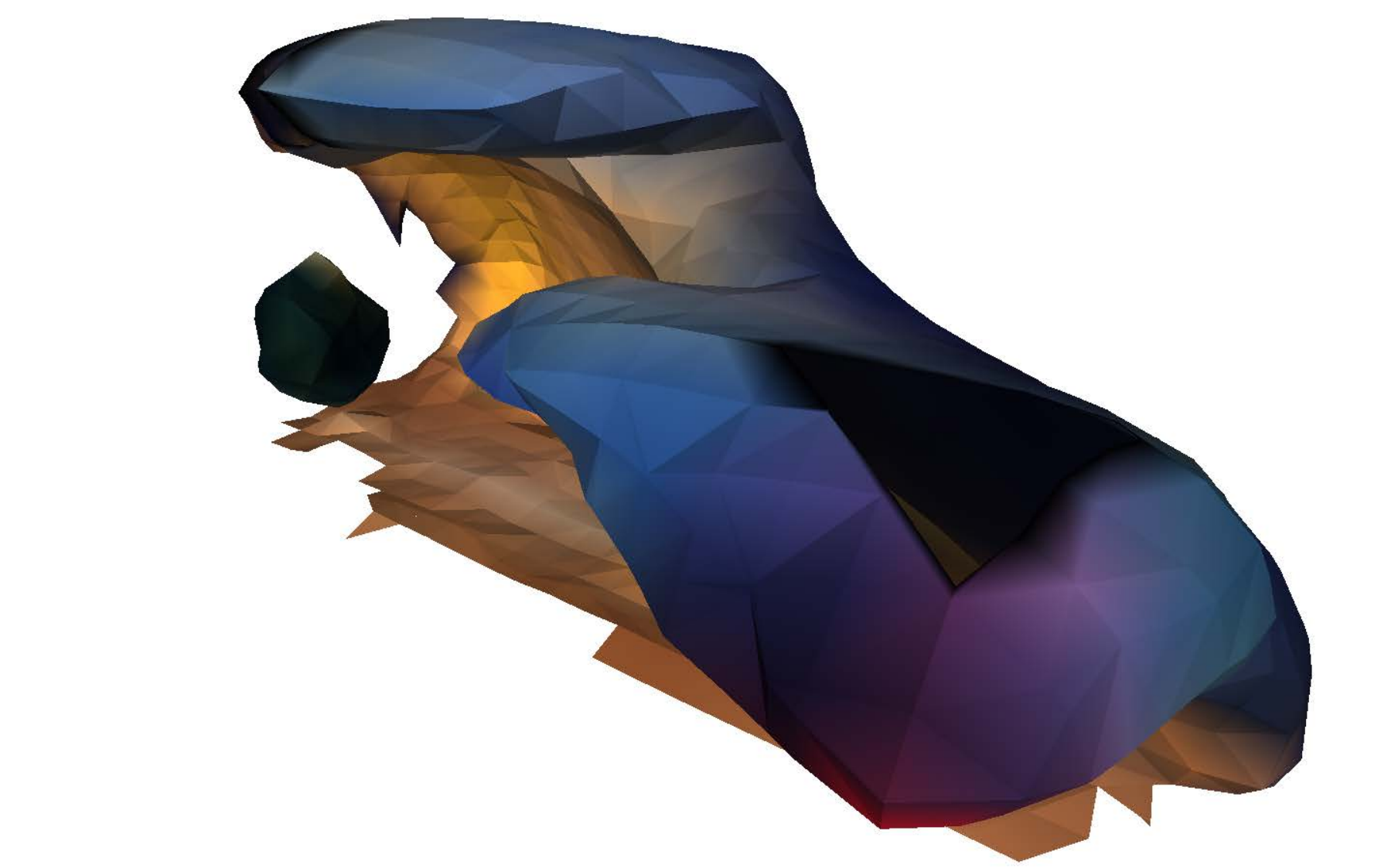}} & 
    \makecell{\includegraphics[width=.3\linewidth]{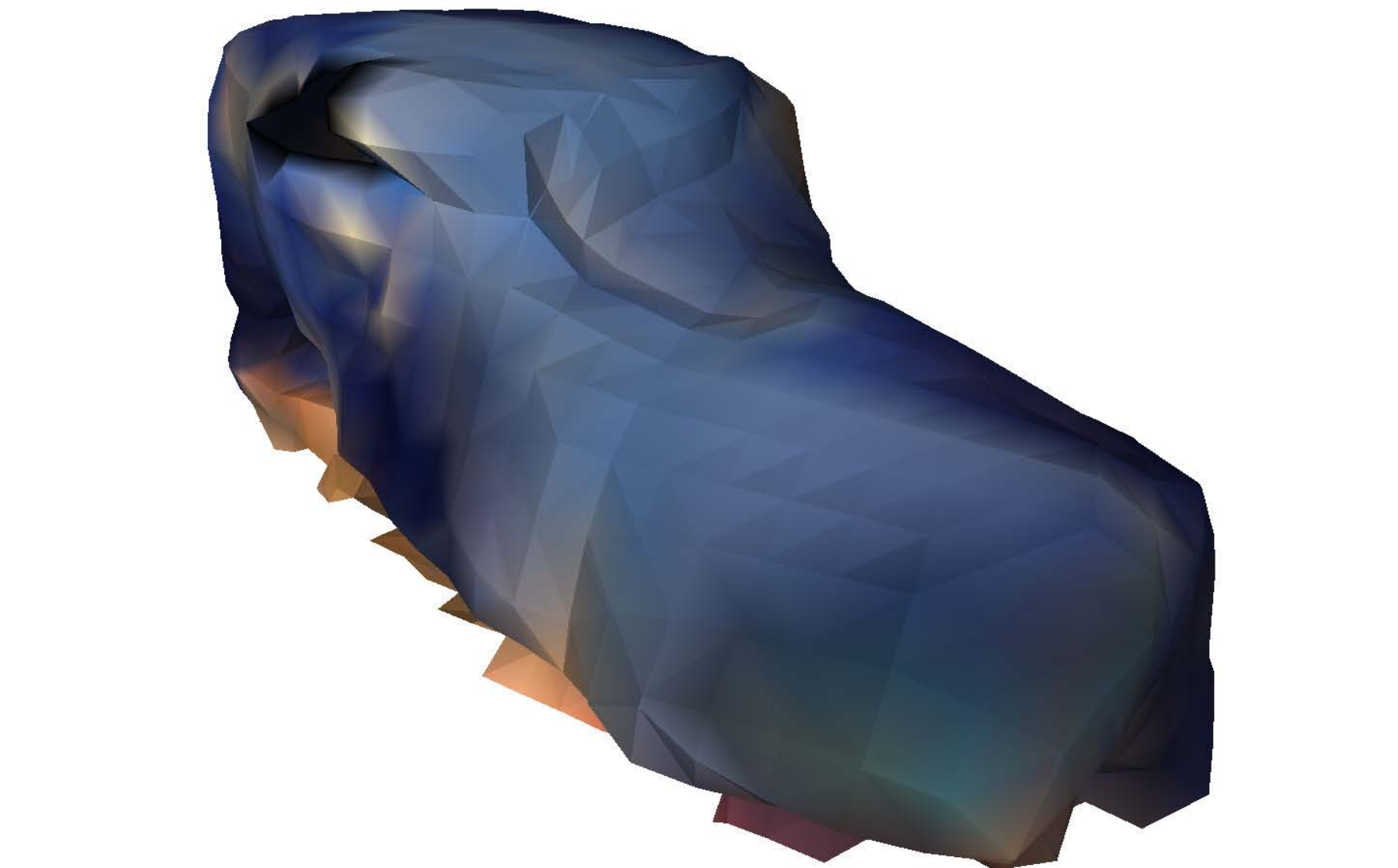}} \vspace{-0.3em}
    
  \end{tabular}
  \caption{\textbf{Demonstration for Objects Completion.} Random Keyframe selection strategy used in Vox-Fusion \cite{yang2022vox} is unable to reconstruct the 3D mesh that does not appear in the current view (Left column). Ours, based on overlap, however, fully reconstructs the blue car (Right column), even for some parts of the car that are not observed in the current camera view.}
    \label{fig:novel}
    \vspace{-0.3cm}
\end{figure}

We visualize the reconstructed results of a blue car of two different keyframe-selection strategies in Fig. \ref{fig:novel}. It is evident that random-selection strategy generates more gaps on the hood of the blue car, and the occluded area in the direction of the camera view is empty. However, the generative ability is out of the scope of this paper, as the part never observed by the camera is not completely filled by either strategy.

\subsection{Ablation Study of Time and Mask}

\begin{figure}[ht!]
  \centering
  \setlength{\tabcolsep}{0.7pt}

  \begin{tabular}{ccc}
    \multicolumn{1}{c}{ w/o mask} & \multicolumn{1}{c}{ w/o time} & \multicolumn{1}{c}{ ours} \vspace{-0.1em} \\
     \makecell{\includegraphics[width=.32\linewidth]{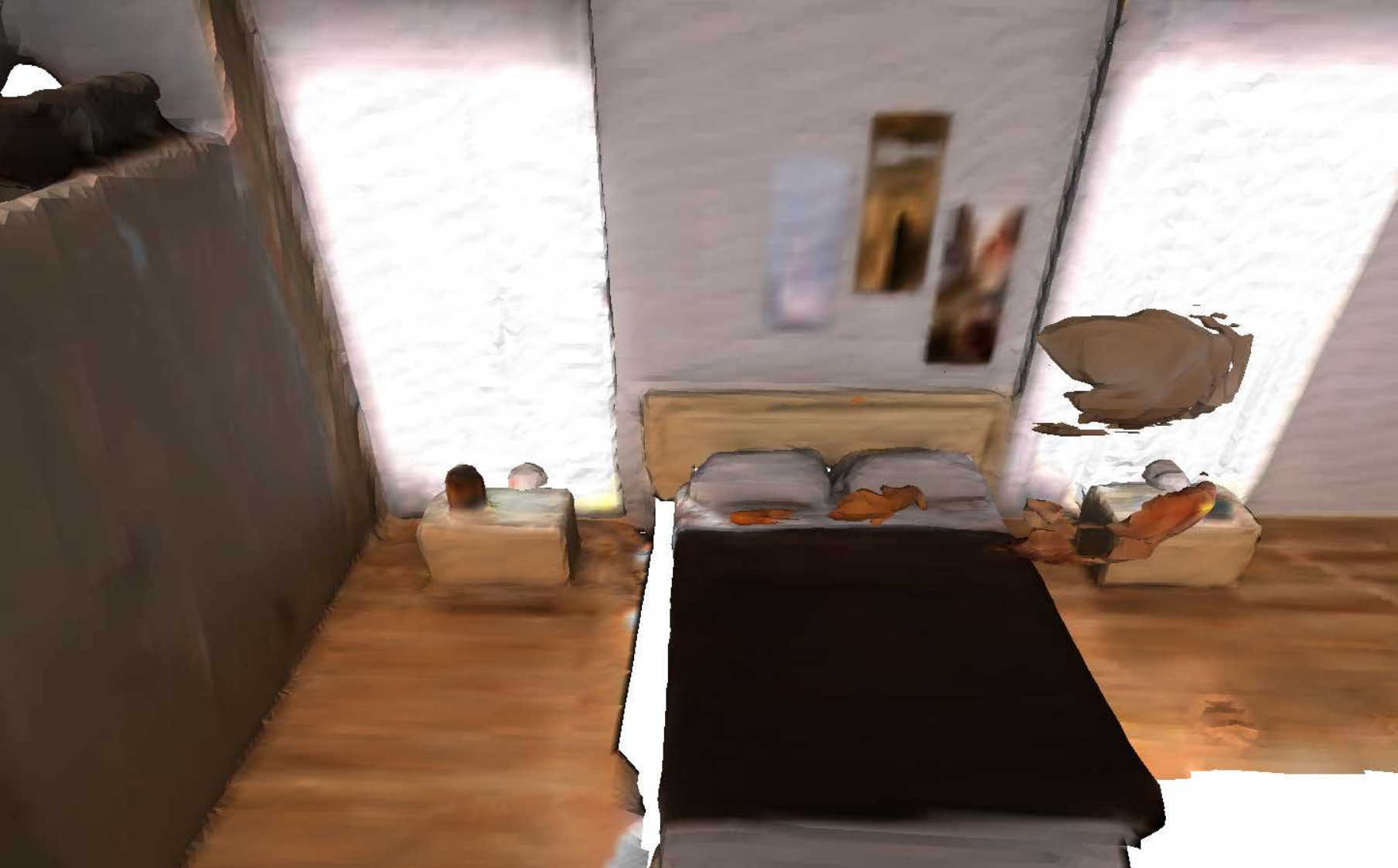}} & 
    \makecell{\includegraphics[width=.32\linewidth]{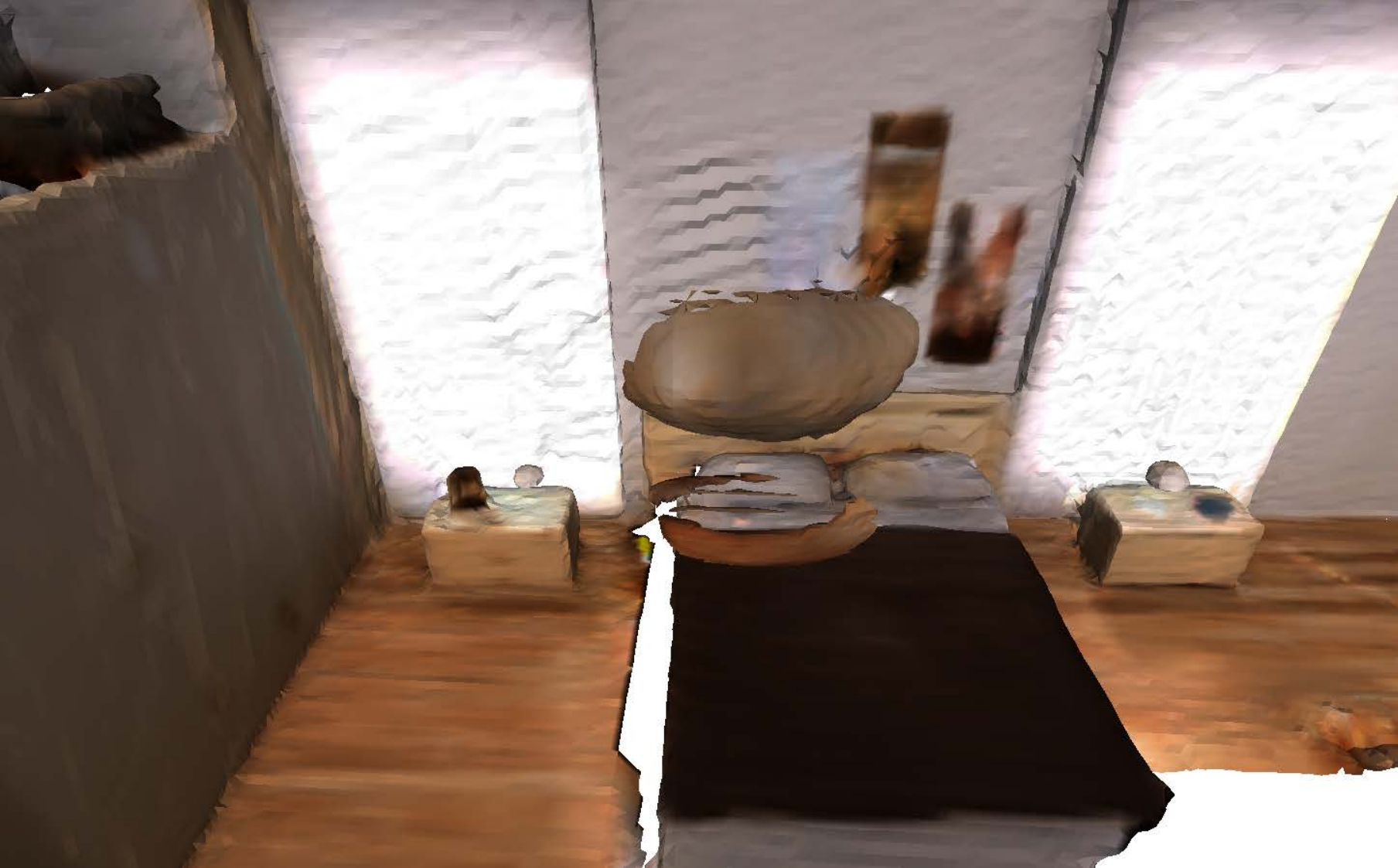}} &
    \makecell{\includegraphics[width=.32\linewidth]{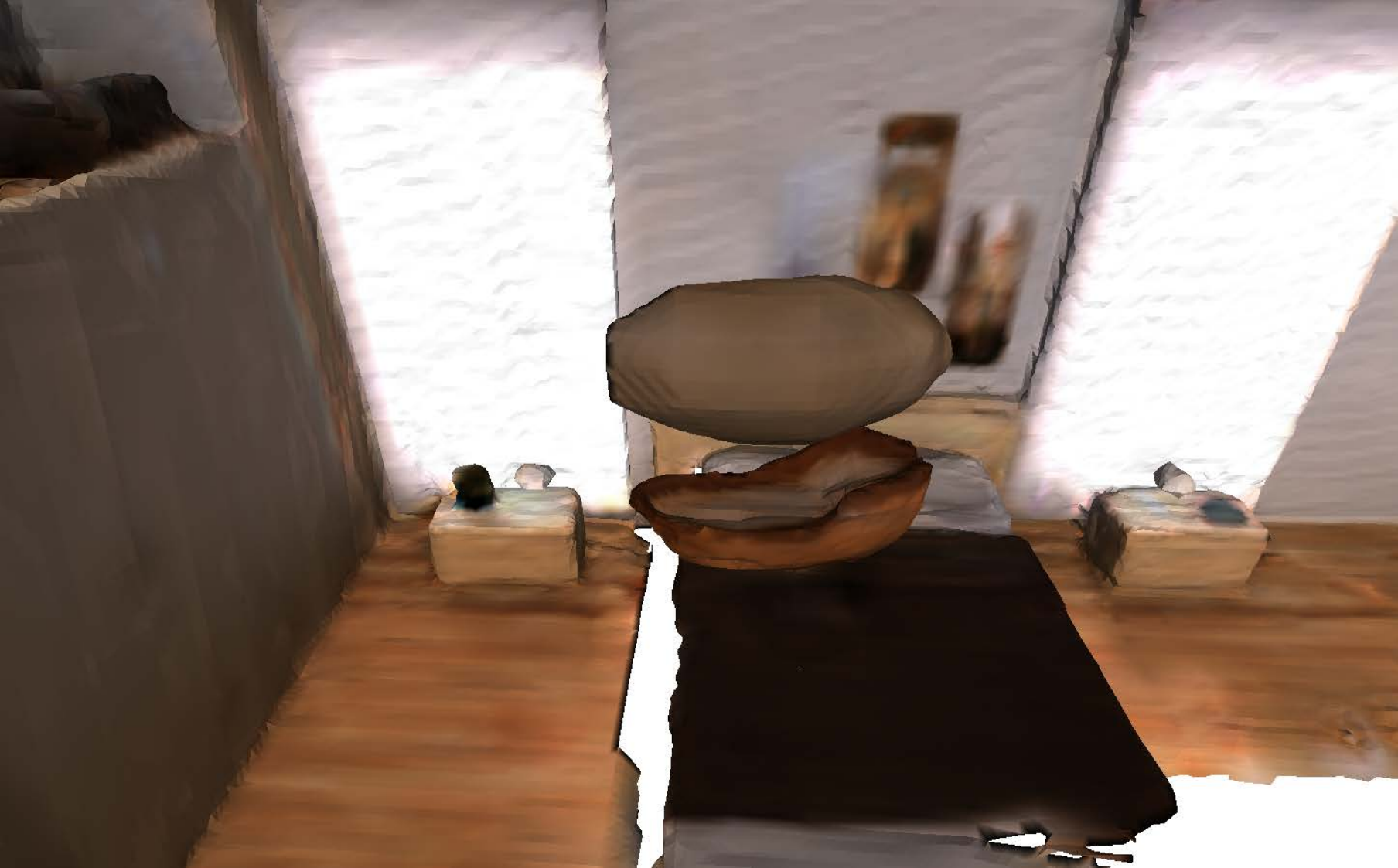}}\vspace{-3pt} \\
    \makecell{\includegraphics[width=.32\linewidth]{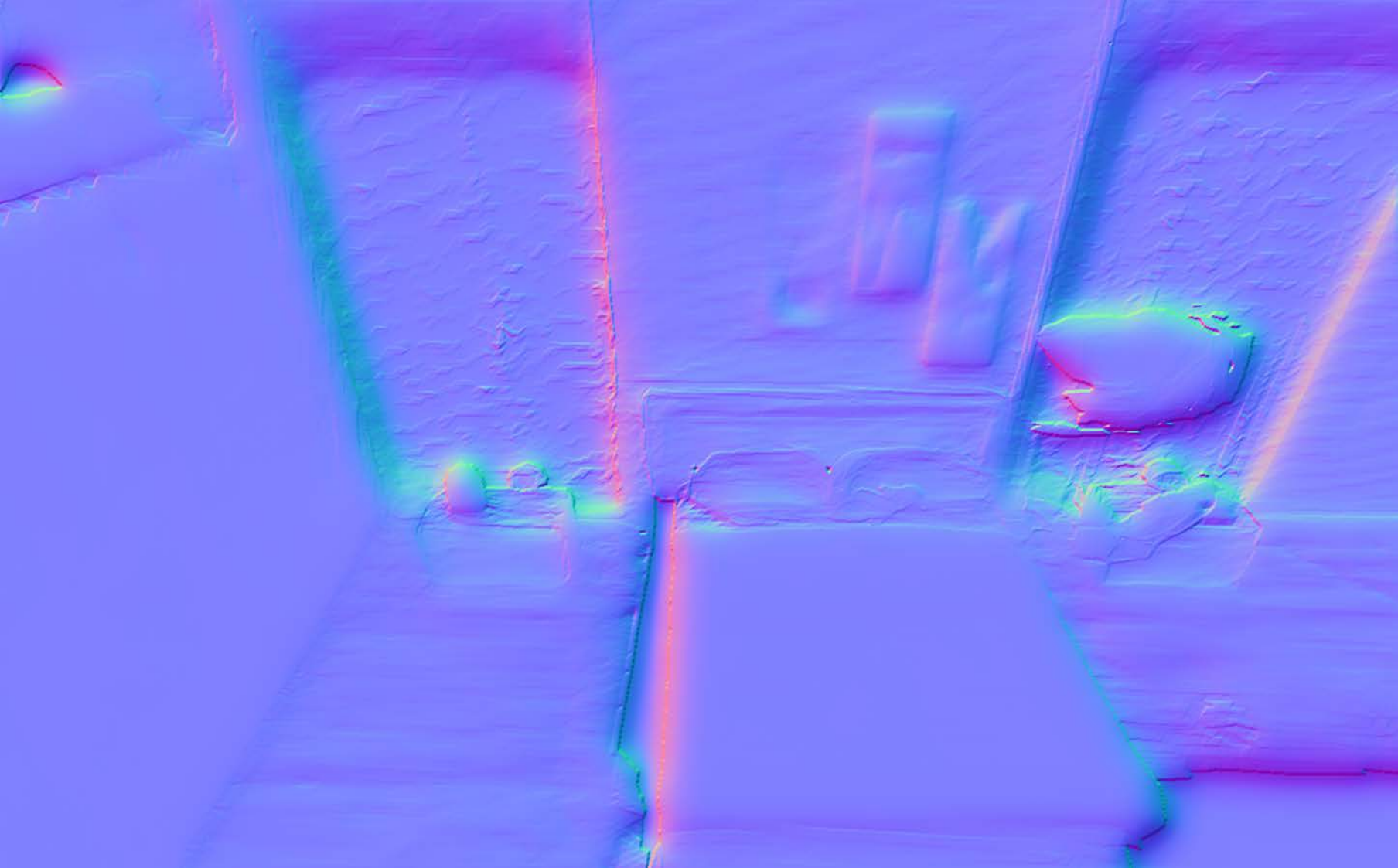}} & 
    \makecell{\includegraphics[width=.32\linewidth]{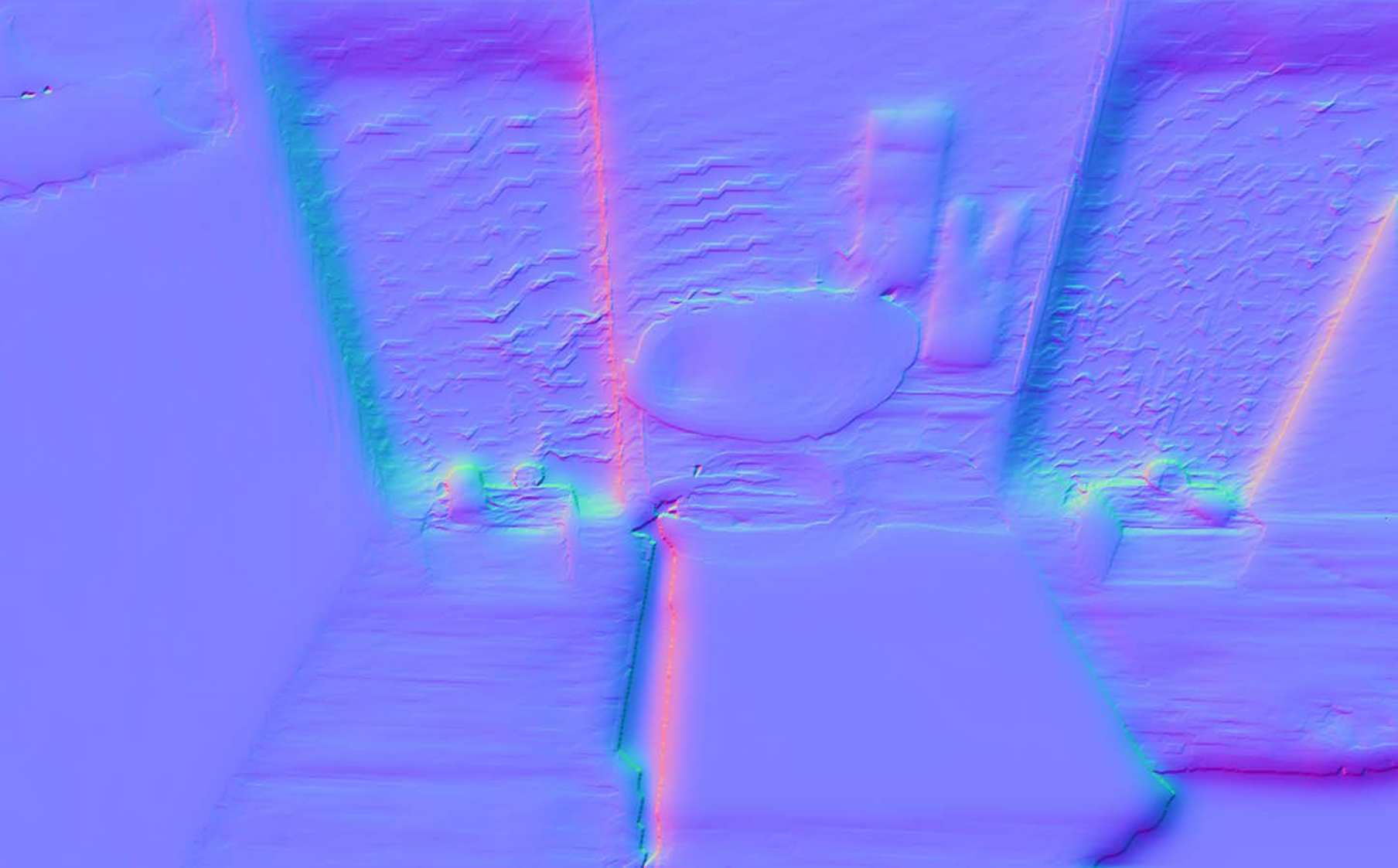}} &
    \makecell{\includegraphics[width=.32\linewidth]{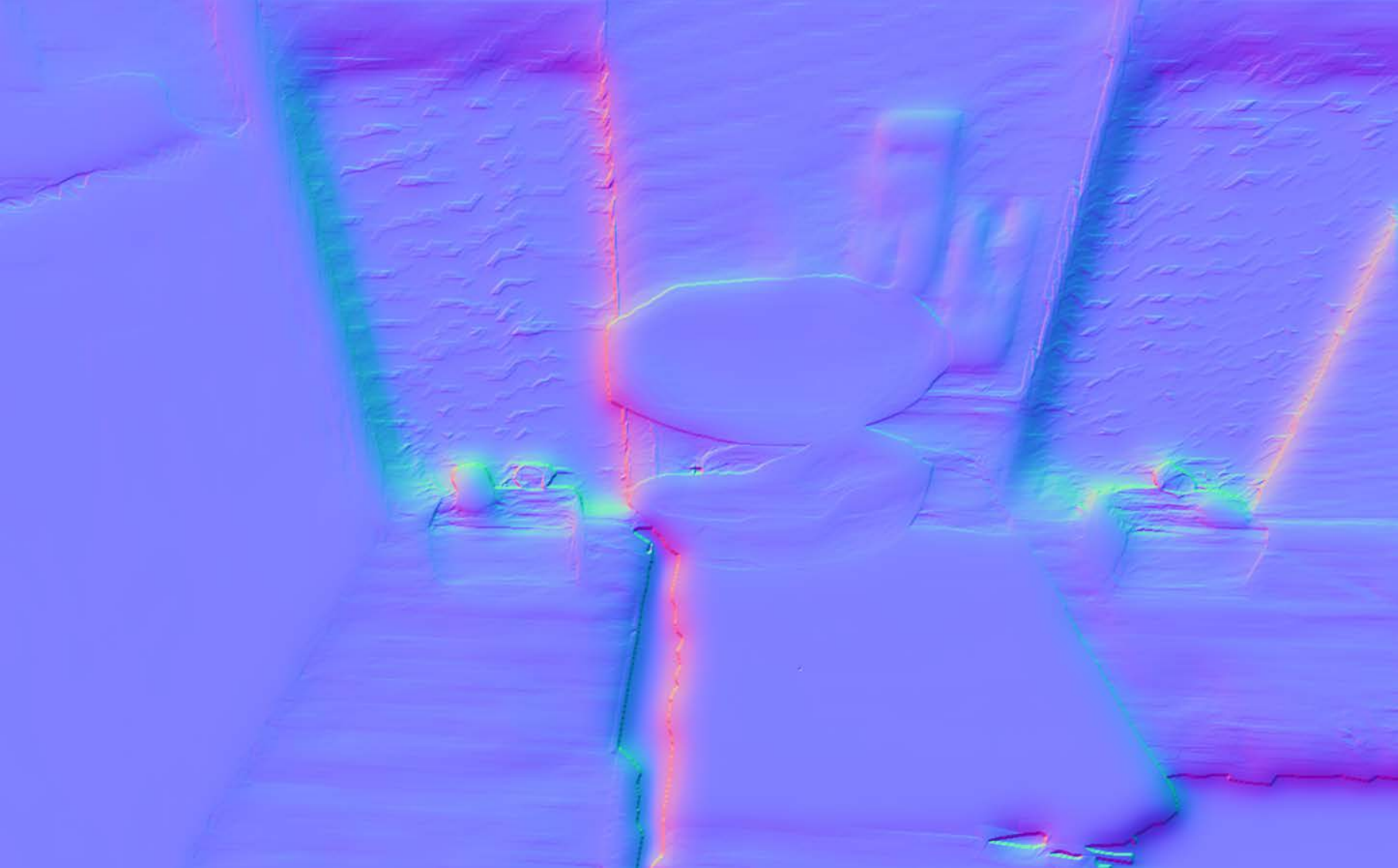}}\vspace{-2pt} 
  \end{tabular}
   \caption{\textbf{Ablation Study for Time and Mask.} The rendered images and their Norm images excluding the dynamic masks and the time $t$ are reported.  Rendering is executed with the $609_{th}$ frame of Dataset Room4. \textbf{w/o mask:} Reconstructed flying ship is incomplete and registered at incorrect positions. \textbf{w/o time:} Occasionally producing intermediate results, there are many holes on the ship and distorted lines on the wall.}
    \label{fig:ablation}\
    \vspace{-1cm}
\end{figure}

In this paper, we associate 3D positions with the time $t$ and increase the number of sampling points within the area indicated by dynamic masks. To validate the effectiveness of two improvements, Fig. \ref{fig:ablation} and TABLE \ref{table:ablationeva} depict the results from ablation experiments. We conduct experiments where we separately omit masks and  time $t$ of dynamic objects, then compare their results with the proposed approach. It is evident that our method achieves the best results.


\begin{table}[ht]
\caption{\small Quantitative Evaluation for Ablation Study}
\label{table:ablationeva}
\begin{center}
\begin{tabular}{c||c||>
{\centering\arraybackslash}p{0.85cm}|>{\centering\arraybackslash}p{0.85cm}|>{\centering\arraybackslash}p{0.85cm}|>{\centering\arraybackslash}p{0.85cm}}
\hline
Datasets& Methods & MSE($\downarrow$)& PSNR($\uparrow$)&SSIM($\uparrow$)&LPIPS($\downarrow$)\\
\hline
\multirow{3}{*}{\makecell{Room4}} 
&w/o mask & 0.0051 & 23.2877& 0.7070& 0.4384\\ 
&w/o time & \second0.0027 & \second26.2984 & \second0.7909&\second0.3948\\
&ours & \fst0.0020 & \fst27.2393 &\fst0.8103& \fst0.3711\\

\hline
\multirow{3}{*}{\makecell{ToyCar3}} 
&w/o mask & \second0.0033 & \second24.9768& 0.8868& 0.1884\\ 
&w/o time & \second0.0033 & 24.9712& \second0.9173 & \second0.1767\\
&ours & \fst0.0032 & \fst25.3456 & \fst0.9213 & \fst0.1604\\
\hline
\end{tabular}
\end{center}
\vspace{-0.5cm}
\end{table}

\section{CONCLUSION}\label{sec:conclusion}
This paper introduces a novel dense SLAM system that leverages time-varying neural implicit representations to understand dynamic scenes. Our novelty lies in the introduction of a time-varying representation to extend 3D space positions to 4D space-temporal positions. Additionally, we introduce a more efficient keyframe selection strategy by calculating the overlapping ratio between the current frame and each frame in the keyframes database. Additionally, our keyframe selection strategy enables us to construct more complete dynamic objects. As opposed to existing dynamic SLAMs, our method does not rely on pre-trained models to reconstruct the dynamic objects. 

\textbf{Limits \& Future Work:}
The trade-off between mapping quality of real-world scenes and complexity of DNNs, as well as camera pose estimation with high-speed moving objects are potential research directions for our future endeavors.




\bibliographystyle{IEEEtran}
\bibliography{10_mybib}
\end{document}